\documentclass[sigconf]{aamas} 

\usepackage{balance} 
\usepackage{float}
\usepackage{tabularx}
\usepackage{caption}
\usepackage{subcaption}

\setcopyright{none}
\acmConference[ALA '22]{Proc.\@ of the Adaptive and Learning Agents Workshop (ALA 2022)}
{May 9-10, 2022}{Online, \url{https://ala2022.github.io/}}{Cruz, Hayes, da Silva, Santos (eds.)}
\copyrightyear{2022}
\acmYear{2022}
\acmDOI{}
\acmPrice{}
\acmISBN{}
\acmSubmissionID{23}

\title   [An Analysis of Discretization Methods for Communication Learning with Multi-Agent Reinforcement Learning]{An Analysis of Discretization Methods for Communication Learning with Multi-Agent Reinforcement Learning}

\author{Astrid Vanneste}
\affiliation{
  \institution{University of Antwerp - imec \\
               IDLab - Faculty of Applied Engineering}
  \city{Antwerp}
  \country{Belgium}}
\email{astrid.vanneste@uantwerpen.be}

\author{Simon Vanneste}
\affiliation{
  \institution{University of Antwerp - imec \\
               IDLab - Faculty of Applied Engineering}
  \city{Antwerp}
  \country{Belgium}}
\email{simon.vanneste@uantwerpen.be}

\author{Kevin Mets}
\affiliation{
  \institution{University of Antwerp - imec\\
              IDLab - Department of Computer Science}
  \city{Antwerp}
  \country{Belgium}}
\email{kevin.mets@uantwerpen.be}

\author{Tom De Schepper}
\affiliation{
  \institution{University of Antwerp - imec\\
              IDLab - Department of Computer Science}
  \city{Antwerp}
  \country{Belgium}}
\email{tom.deschepper@uantwerpen.be}

\author{Siegfried Mercelis}
\affiliation{
  \institution{University of Antwerp - imec \\
               IDLab - Faculty of Applied Engineering}
  \city{Antwerp}
  \country{Belgium}}
\email{siegfried.mercelis@uantwerpen.be}

\author{Steven Latré}
\affiliation{
  \institution{University of Antwerp - imec\\
              IDLab - Department of Computer Science}
  \city{Antwerp}
  \country{Belgium}}
\email{steven.latre@uantwerpen.be}

\author{Peter Hellinckx}
\affiliation{
  \institution{University of Antwerp - imec \\
               IDLab - Faculty of Applied Engineering}
  \city{Antwerp}
  \country{Belgium}}
\email{peter.hellinckx@uantwerpen.be}

\newcommand{\graphwidth}{0.45}
\newcommand{\confusionwidth}{0.37}
\newcommand{\histogramwidth}{0.9}

\begin{abstract}
Communication is crucial in multi-agent reinforcement learning when agents are not able to observe the full state of the environment. The most common approach to allow learned communication between agents is the use of a differentiable communication channel that allows gradients to flow between agents as a form of feedback. However, this is challenging when we want to use discrete messages to reduce the message size since gradients cannot flow through a discrete communication channel. Previous work proposed methods to deal with this problem. However, these methods are tested in different communication learning architectures and environments, making it hard to compare them. In this paper, we compare several state-of-the-art discretization methods as well as two methods that have not been used for communication learning before. We do this comparison in the context of communication learning using gradients from other agents and perform tests on several environments. Our results show that none of the methods is best in all environments. The best choice in discretization method greatly depends on the environment. However, the discretize regularize unit (DRU), straight through DRU and the straight through gumbel softmax show the most consistent results across all the tested environments. Therefore, these methods prove to be the best choice for general use while the straight through estimator and the gumbel softmax may provide better results in specific environments but fail completely in others.
\end{abstract}


\begin{document}


\pagestyle{fancy}
\fancyhead{}


\maketitle 


\begin{table*}[ht]
\begin{minipage}{\textwidth}
\centering
\caption{Comparison the state of the art and our work \\  {\normalfont (DRU - Discretize Regulize Unit, GS - Gumbel Softmax, STE - Straight Through Estimator, ST-DRU - Straight Through DRU, ST-GS - Straight Through GS)}}
\label{tab:sota}
\begin{tabularx}{\textwidth}{llXl}
\toprule
                                                    & Message Type          & Communication Learning Technique & Discretization Method  \\
                                            \midrule
RIAL \citep{foerster2016learning}                   &  Discrete             & DQN using team reward                 &  Discrete policy                  \\
DIAL \citep{foerster2016learning}                   &  Discrete             & Gradients from other agents           &  DRU                  \\
CommNet \citep{sukhbaatar2016learning}, A3C3 \citep{simoes2020a3c3}              &  Continuous           & Gradients from other agents           &  N/A                  \\
MADDPG \citep{lowe2020multiagent, mordatch2018}     &  Continuous/Discrete  & Gradients from the critic             &  GS       \\
\citet{Freed_Sartoretti_Hu_Choset_2020}             &  Discrete             & Gradients from other agents           &  Randomized Encoder/Decoder \\
\citet{jaques2019socialinfluence}                   &  Discrete             & A3C using team reward augmented with social influence reward              &  Discrete policy                     \\
MACC \citep{vanneste2021learning}                   &  Discrete             & Counterfactual reasoning              &  Discrete policy                     \\
\citet{lin2021learning}                             &  Discrete             & Reconstruction loss                   &  STE                     \\
\textbf{This Work}                                  &  \textbf{Discrete}    & \textbf{Gradients from other agents}  &  \textbf{(ST)-DRU / STE / (ST)-GS}                     \\
\bottomrule
\end{tabularx}
\end{minipage}
\end{table*}

\section{Introduction}
Over the past several years, both single-agent reinforcement learning (RL) and multi-agent RL (MARL) have gained a lot of interest. In (MA)RL, the agents often have to deal with partial observability. The agents can only observe part of the global state, making it hard to choose appropriate actions. For example, an agent navigating through an environment cannot see certain parts of the environment. In MARL, this partial observability can often be alleviated by allowing the agents to share information with each other. By combining this information with their own observation, agents get a more complete view of the environment and can choose better actions \citep{tan1993multi, melo2012}. For example, multiple agents navigating through the same environment can share the information they can see with each other, resulting in a more complete view. 

One of the subfields within MARL is research towards learned communication between agents. The most commonly used approach thus far is allowing gradients to flow between agents as a form of feedback on the received messages. However, in the case of discrete communication messages this raises a problem since gradients cannot flow through a discrete communication channel. Several different approaches have been proposed in the state of the art \citep{foerster2016learning, lowe2020multiagent, mordatch2018, lin2021learning} to discretize messages while allowing gradients to flow through the discretization unit. Each of these methods was tested using different communication learning approaches and applied on different environments, making a fair comparison very difficult. 

Our contributions consist of two parts. First, we present an in-depth comparison of several discretization methods used in the state-of-the-art. In our comparison we focus on using these discretization methods to allow discrete communication when learning communication using the gradients of the receiving agents. We compare each of the approaches on several environments with increasing complexity as well as analyze their performance when the environment introduces errors to the communication messages. Secondly, we present two discretization methods (ST-DRU and ST-GS) that have not been used in communication learning before.  

The remainder of this paper is structured as follows. Section \ref{sec:related_work} gives an overview of work related to our research. Section \ref{sec:background} contains some additional background information. Section \ref{sec:methods} provides a detailed explanation of the discretization methods we compare in this paper. In Section \ref{sec:experiments}, the different experiments that were done are explained along with their results. We discuss the results of our experiments further in Secion \ref{sec:discussion}. In Section \ref{sec:conclusion} we draw some conclusions from our experimental results.

\section{Related Work}
\label{sec:related_work}
In this section, we review state-of-the-art work relevant to our research. We give an overview of several communication learning methods, focusing on methods that learn discrete communication. Here, we see some alternative approaches for learning discrete communication beside using a differentiable communication channel as well as different discretization techniques used in the state of the art. 

\citet{foerster2016learning} and \citet{sukhbaatar2016learning} proposed the first successful methods for learning inter-agent communication. \citet{foerster2016learning} proposed two novel approaches, Reinforced Inter-Agent Learning (RIAL) and Differentiable Inter-Agent Learning (DIAL). Both RIAL and DIAL use discrete communication messages, but learn the communication policy in a different way. RIAL learns the communication policy the same way as learning the action policy, by using the team reward. However, the results clearly show that this is not sufficient in most environments. DIAL proved more successful by using gradients originating from the agents receiving the messages which provide feedback on the communication policy. \citet{sukhbaatar2016learning} proposed a different approach called CommNet. Messages consist of the hidden state of the agents, resulting in continuous messages. Similar to DIAL, CommNet uses gradients that flow through the communication channel to train the communication. 

\begin{figure*}[ht]
    \centering
    \includegraphics[width=0.6\textwidth]{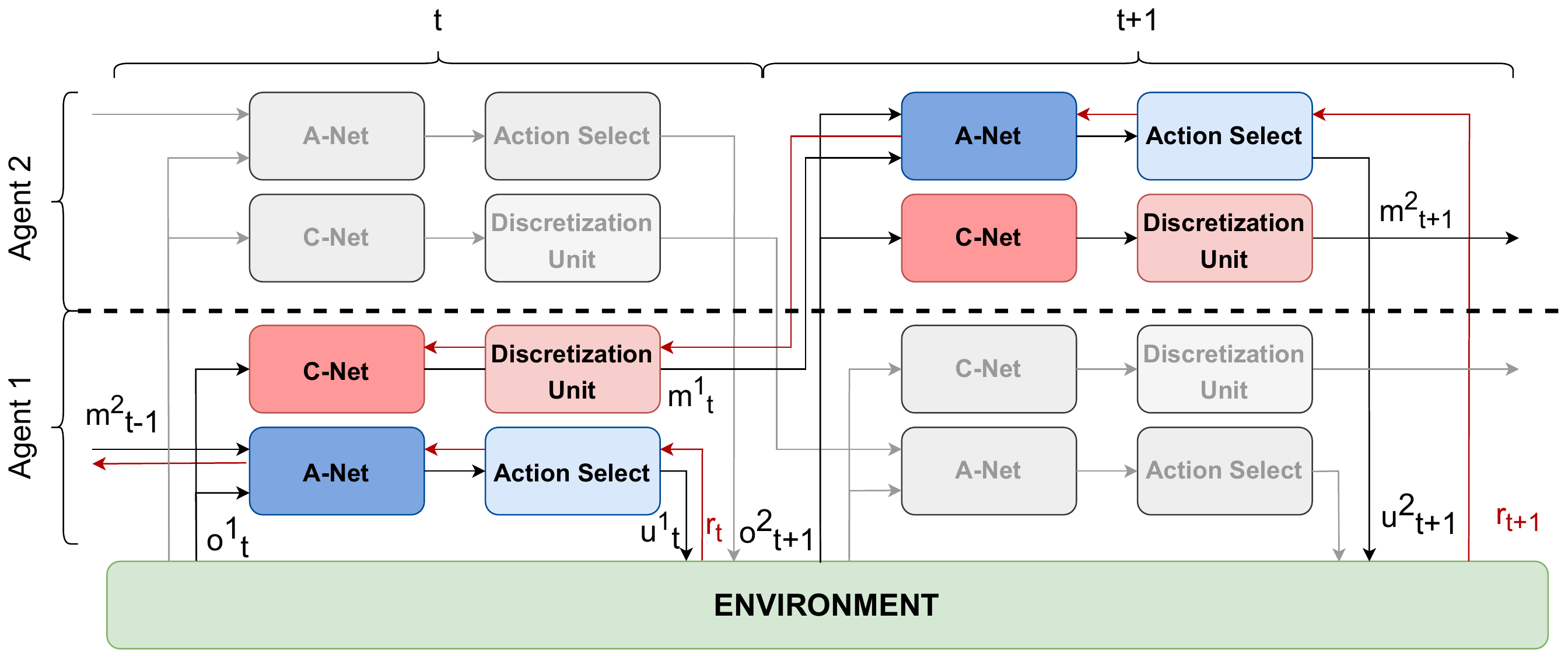}
    \caption{Architecture of DIAL}
    \label{fig:DIAL}
    \Description{Figure 1. Fully described in text}
\end{figure*}

A lot of the research that follows these works chooses to use continuous communication like \citet{sukhbaatar2016learning}, avoiding the problem of discretizing the communication messages \citep{simoes2020a3c3}. 
Other methods choose a different method to learn communication than using gradients through the communication channel. This also avoids the challenge of discretizing the messages.
\citet{jaques2019socialinfluence} train the communication policy using the team reward augmented with a social influence reward. This additional reward is based on how much the message changes the action policy of the receiving agents. 
\citet{vanneste2021learning} use counterfactual reasoning to directly learn a communication protocol without the need of a differentiable communication channel. 
\citet{Freed_Sartoretti_Hu_Choset_2020} use a randomized encoder at the sender to encode the continuous messages into discrete messages. At the receiver, a randomized decoder is used to approximate the original continuous message. They show that by using this technique they can consider the communication channel equivalent to a continuous channel with additive noise, allowing gradients to flow between the sender and receiver agent.

\citet{lowe2020multiagent} and \citet{mordatch2018} propose Multi-Agent Deep Deterministic Policy Gradients (MADDPG). In their work they evaluate MADDPG on multiple different scenarios, including communication tasks. They do not use a differentiable communication channel to learn communication but they have to make sure the messages are differentiable to allow the MADDPG method to work properly since policies are learned using gradients that originate from the critic. They allow discrete communication by using a gumbel softmax. 
\citet{lin2021learning} use an autoencoder at the sender to compose a representation of the observation that will be used as communication message. To discretize these messages they use a straight through estimator in the autoencoder. 
Both \citet{lowe2020multiagent, mordatch2018} and \citet{lin2021learning} have to use differentiable discretization techniques in their methods to allow discrete communication. However, they do not use the techniques in the same way we do in our work. \citet{lowe2020multiagent} and \citet{mordatch2018} use the discretization method in a similar way as our work but in MADDPG the gradients that correct the communication originate from the critic instead of from other agents. \citet{lin2021learning} use the discretization method in a very different way since they train the communication policy entirely using the reconstruction loss of the autoencoder instead of the gradients from the other agents.  

Summarized, in the state-of-the-art related to our research, we see that multiple discretization methods have been proposed. But, differences in communication learning approaches and the fact that each of these methods is tested on different environments makes comparing these different discretization methods very hard. 


\section{Background}
\label{sec:background}

\subsection{Deep Q-Networks (DQN)}
In single agent RL \citep{sutton1998introduction}, the agent chooses an action $u_t \in U$ based on the state $s_t \in S$ of the environment. As a result of this action, the environment will transition to a new state $s_{t + 1}$ and provide the agent with a reward $r_{t+1} \in R$. This reward is used to train the agent. Q-learning uses this reward to calculate a Q-value for each state action pair $Q(s, u)$. This Q-value represents a value for each state action pair, where a higher Q-value indicates a better action. Therefore, the policy of our agent can be defined by Equation \ref{eq:qlearning_policy}.

\begin{equation}
    \begin{aligned}
        \pi(s) = \underset{u}{argmax} \left( Q(s, u) \right)
    \end{aligned}
    \label{eq:qlearning_policy}
\end{equation}

Deep Q-learning \citep{mnih2015human} uses a neural network with parameters $\theta$ to represent the Q-function. The deep Q-network is optimized at iteration $i$ by minimizing the loss in Equation \ref{eq:dqn_loss}.

\begin{equation}
    \begin{aligned}
        \mathcal{L}_i(\theta_i) = \mathbb{E}_{s_t,u_t,r_t,s_{t+1}} \big[ (r_t + \gamma \max_{u_{t}} Q(s_{t+1},u_{t+1}, \theta^-_i) - Q(s_t,u_t, \theta_i))^2\big]
    \end{aligned}
    \label{eq:dqn_loss}
\end{equation}
where $\gamma$ is the discount factor and $\theta^-_i$ are the parameters of the target network. This target network will be updated after each training iteration according to Equation \ref{eq:target_network_update}.
\begin{equation}
    \begin{aligned}
        \theta^-_{i + 1} \leftarrow \tau \theta_i + (1 - \tau) \theta^-_i
    \end{aligned}
    \label{eq:target_network_update}
\end{equation}
where $\tau$ is a weight that indicates how fast the target network should follow the parameters $\theta$.
In our work, the agent does not receive the full state $s_t$ but only a limited observation $o_t \in O$ of this state. This increases the complexity since the observation might lack important information.

\begin{table*}[ht]
\caption{Differences between the discretization methods where $x$ is the input of the discretization unit (the output of the C-Net), $H(x)$ is the heaviside function, $n \sim \mathcal{N}(0, \sigma_G^2)$ is Gaussian noise, $i \in \{0, 1\}$, $g_i \sim G$ is Gumbel noise, $\pi_0 = \sigma(x)$, $\pi_1 = (1 - \sigma(x))$, $\tau$ is the softmax temperature and $\sigma(x)$ is the sigmoid function} 
\label{tab:discretization_units}
\begin{tabular}{cccc}
\toprule
                & Training Output (forward pass)            & Function used for backward pass  & Evaluation Output (forward pass) \\ \midrule
STE             & $H(x)$                                    & $x $                      & $H(x)$            \\ 
DRU             & $\sigma(x + n)$                          & $\sigma(x + n)$          & $H(x)$            \\ 
GS              & $softmax \left(\frac{log(\sigma(x)) + g_1}{\tau}, \frac{log(1-\sigma(x)) + g_2}{\tau} \right)[0]$                          & $softmax \left(\frac{log(\sigma(x)) + g_1}{\tau}, \frac{log(1-\sigma(x)) + g_2}{\tau} \right)[0]$                                  & $one\_hot(\underset{i}{argmax}[g_i + log \pi_i])[0]$                  \\
ST-DRU          & $H(x + n)$                                & $\sigma(x + n)$           & $H(x)$            \\
ST-GS           & $one\_hot(\underset{i}{argmax}[g_i + log \pi_i])[0]$             & $softmax \left(\frac{log(\sigma(x)) + g_1}{\tau}, \frac{log(1-\sigma(x)) + g_2}{\tau} \right)[0]$                                  & $one\_hot(\underset{i}{argmax}[g_i + log \pi_i])[0]$                   \\
\bottomrule
\end{tabular}
\end{table*}

\begin{figure*}
    \begin{minipage}[t]{0.49\linewidth}
        \centering
        \includegraphics[width=\histogramwidth\linewidth, trim=65 10 80 35, clip]{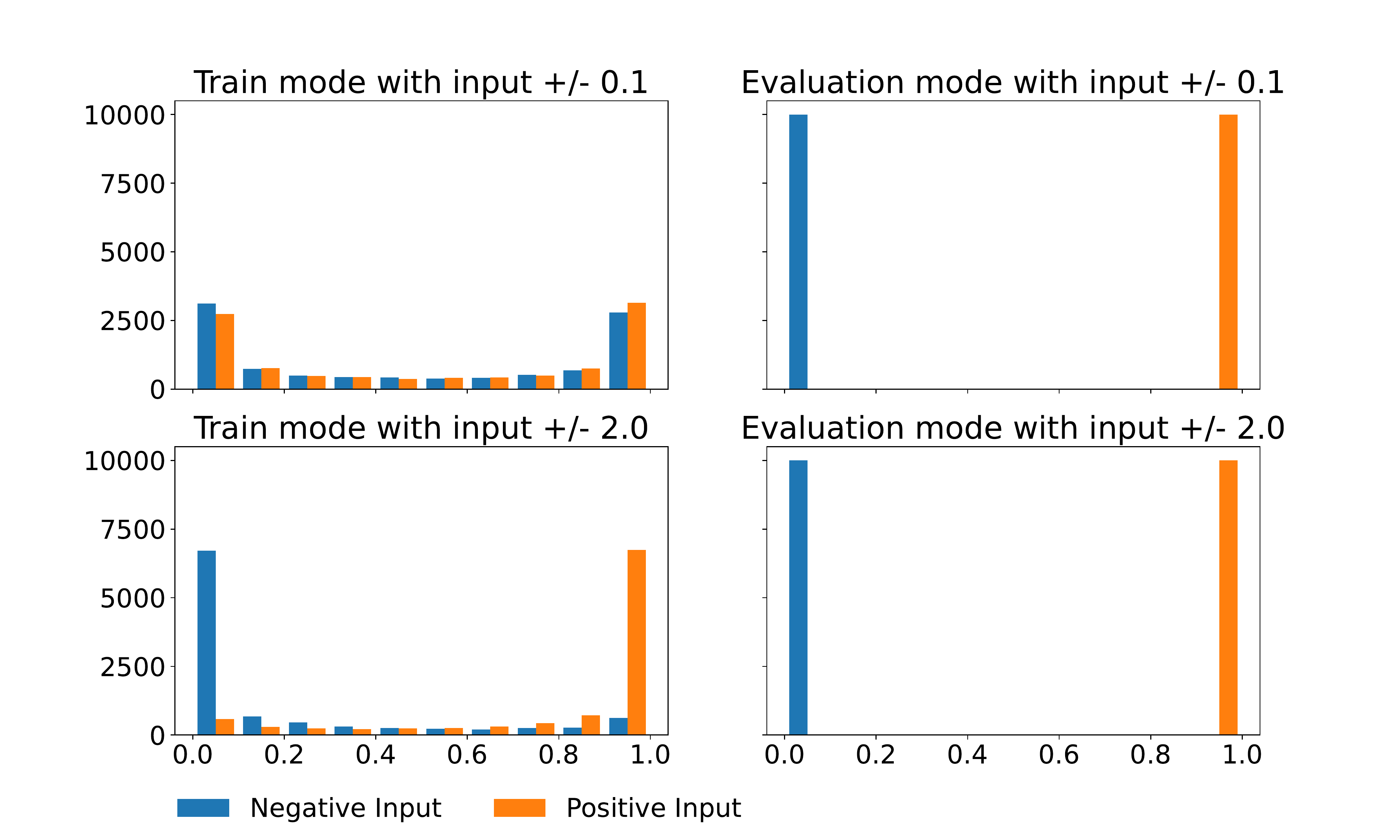}
        \caption{Histogram of the output of the DRU when calculating the output 10k times for different input values (positive and negative) for both train and evaluation mode}
        \label{fig:DRU_response}
        \Description[Figure 2. Histogram of the output of the Discretize Regularize Unit]{The figure shows four histograms. On each histogram, we see the output of the Discretize Regularize Unit for 10000 times a positive input and 10000 times a negative input. On the left we see the output when the Discretize Regularize Unit is in training mode and on the right we see the output when the Discretize Regularize Unit is in evaluation mode. The top left histogram shows the outputs of the Discretize Regularize Unit in train mode for an input of plus and minus 0.1. We see that the outputs for the positive and negative input overlaps a lot. We see that a little over 25\% of the outputs of both the positive and negative inputs results in an output of 0 and 1. The other inputs are resulting in outputs between 0 and 1 with almost the same number of cases for each output value but slightly more cases towards 0 and 1. On the histogram on the bottom left we see the outputs of the Discretize Regularize Unit for an input of plus and minus 2.0. We see that there is a lot less overlap between the positive and negative input. Around 60\% of the cases of the negative input results in an output of 0 and around 60\% of the cases of the positive input results in an output of 1. The remaining cases result in an output between 0 and 1 with the outputs of the positive and negative input still overlapping. On the top and bottom right histograms we see the output of the Discretize Regularize Unit in evaluation mode for an input of plus and minus 0.1 and plus or minus 2.0 respectively. Here, we see that all of the negative inputs result in an output of 0 and all of the positive outputs result in an output of 1.}
    \end{minipage}
    \hfill
    \begin{minipage}[t]{0.49\linewidth}
        \includegraphics[width=\histogramwidth\linewidth, trim=65 10 80 35, clip]{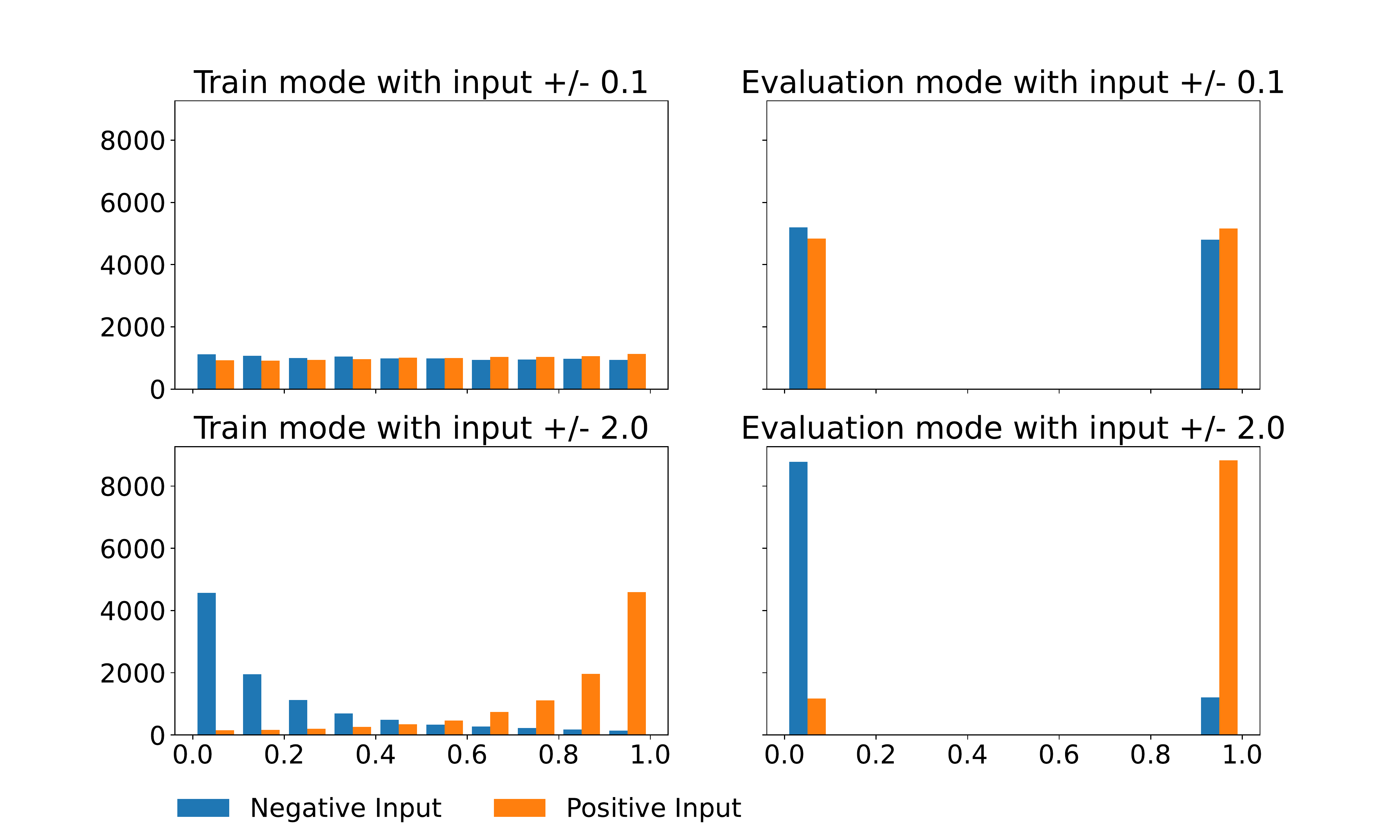}
        \caption{Histogram of the output of the GS when calculating the output 10k times for different input values (positive and negative) for both train and evaluation mode with temparature $\tau = 1.0$}
        \label{fig:GS_response}
        \Description[Figure 3. Histogram of the output of the Gumbel Softmax]{The figure shows four histograms. On each histogram, we see the output of the Gumbel Softmax for 10000 times a positive input and 10000 times a negative input. On the left we see the output when the Gumbel Softmax is in training mode and on the right we see the output when the Gumbel Softmax is in evaluation mode. The top left histogram shows the outputs of the Gumbel Softmax in train mode for an input of plus and minus 0.1. We see that the outputs for both the positive and negative inputs are uniformly distributed between 0 and 1. On the histogram on the bottom left we see the outputs of the Gumbel Softmax for an input of plus and minus 2.0. We see that for the negative inputs a little less than 50\% of the inputs results in an output of 0. The remaining inputs have fewer cases the more the output approaches 1. For the positive inputs we see the same but mirrored around 0.5. A little less than 50\% of the cases results in an output of 1 with the remaining cases resulting in an output between 0 and 1. The amount of cases decreases towards a value of 0. On the top and bottom right histograms we see the output of the Gumbel Softmax in evaluation mode. The top histogram shows the outputs for an input of plus or minus 0.1. We see that approximately 50\% of both the positive and negative inputs results in an output of 0 and the other 50\% results in an output of 1. In the bottom histogram we see the outputs for an input of plus or minus 2.0. We see that over 80\% of the negative inputs results in an output of 0, the remaining cases resulting in an output of 1. For the positive inputs we see that over 80\% of inputs results in an output of 1, the remaining cases resulting in an output of 0.}
    \end{minipage}
\end{figure*}

\subsection{Differentiable Inter-Agent Learning (DIAL)}
To allow for a fair comparison, we will use the same communication learning approach for each of the different discretization methods. We use DIAL as proposed by \citet{foerster2016learning} in our experiments since it is the most general and well known architecture to learn discrete communication using gradients from the other agents. The architecture of DIAL can be seen in Figure \ref{fig:DIAL}. We adapted the original DIAL architecture by separating the action and communication network. This allows us to keep the communication network small, making communication learning easier. In our experiments, we examine environments where the agents only need to share and encode part of the observation which allows us to make this adaptation. When the agents are expected to communicate about a strategy, splitting the action and communication network may no longer be possible.

Each agent consists of two networks, the A-Net and the C-Net. The A-Net produces Q-values to determine the action based on the observation and the incoming messages. The C-Net is responsible for calculating the messages based on the observation. It does not receive the incoming messages in our experiments because in these environments the communication policy does not need the incoming messages to determine the output message. Before the messages are broadcast to the other agents, the discretization unit applies one of the discretization techniques that we are comparing in this paper. To train the agents, we apply the team reward provided by the environment on the A-Net according to deep Q-learning. The gradients from the A-Net are propagated to the C-Net of all the agents that sent a message to that agent. This allows us to train the C-Net using the feedback of the agents receiving the messages.


\section{Methods}
\label{sec:methods}
In this section, we describe the different discretization modules, that we will compare, in more detail. Table \ref{tab:discretization_units} provides an overiew of all the discretization methods and their differences. We show the difference between the function used to calculate the output of the discretization unit during training and during evaluation as well as the function that is used for the backward pass.

\subsection{Discretize Regularize Unit (DRU)}
In the DIAL method, \citet{foerster2016learning} propose a module called the Discretize Regularize Unit (DRU) to allow gradients to be used for training while learning discrete communication messages. The DRU has two modes, discretization and regularization. The discretization mode is used at execution time and discretizes the input into a single bit using Equation \ref{eq:DRU_discretize}.
\begin{equation}
    \begin{aligned}
        m = H(x)
    \end{aligned}
    \label{eq:DRU_discretize}
\end{equation}
where $H(x)$ is a heaviside function and $x$ is the input of the discretization unit (output of the C-Net). This calculation cannot be used during training because the derivative of the heaviside function is the Dirac function which is zero everywhere except at $x=0$, where the output is infinite. Therefore, the regularization mode is used during training. When using the regularization mode, the agents are allowed to communicate using continuous messages. However, the DRU tries to encourage the communication policy to generate messages that can easily be discretized at execution time. This is achieved by applying Equation \ref{eq:DRU_regularize}. 
\begin{equation}
    \begin{aligned}
        m = \sigma(x + n)
    \end{aligned}
    \label{eq:DRU_regularize}
\end{equation}
where $x$ is the input of the discretization unit (output of the C-Net), $n$ is noise sampled from a Gaussian distribution with standard deviation $\sigma_G$ and $\sigma(x)$ is the sigmoid function. The noise will affect the output of the DRU the most when the input is around zero since the sigmoid is the steepest there. The influence will be much smaller for inputs with high absolute values. The output in those cases will also go towards zero and one, making it very similar to discrete, binary messages. This can be seen in Figure \ref{fig:DRU_response}.

\subsection{Straight Through Estimator (STE)}
A straight through estimator (STE) \citep{bengio2013estimating, yin2019understanding} performs a normal discretization, as in Equation \ref{eq:DRU_discretize}, when calculating the output. However, when performing backpropagation, it uses the gradients of an identity function instead of the gradients of the discretization. The advantage of this technique is that the agent receiving the message will immediately receive binary numbers and can learn how to react to these messages while still being able to use the gradients from the receiving agents to train the communication network. For the STE, the output will always look like the DRU in evaluation mode, shown in Figure \ref{fig:DRU_response}.

\begin{figure*}
    \begin{minipage}[t]{0.49\linewidth}
        \centering
        \includegraphics[width=\histogramwidth\linewidth, trim=65 10 80 35, clip]{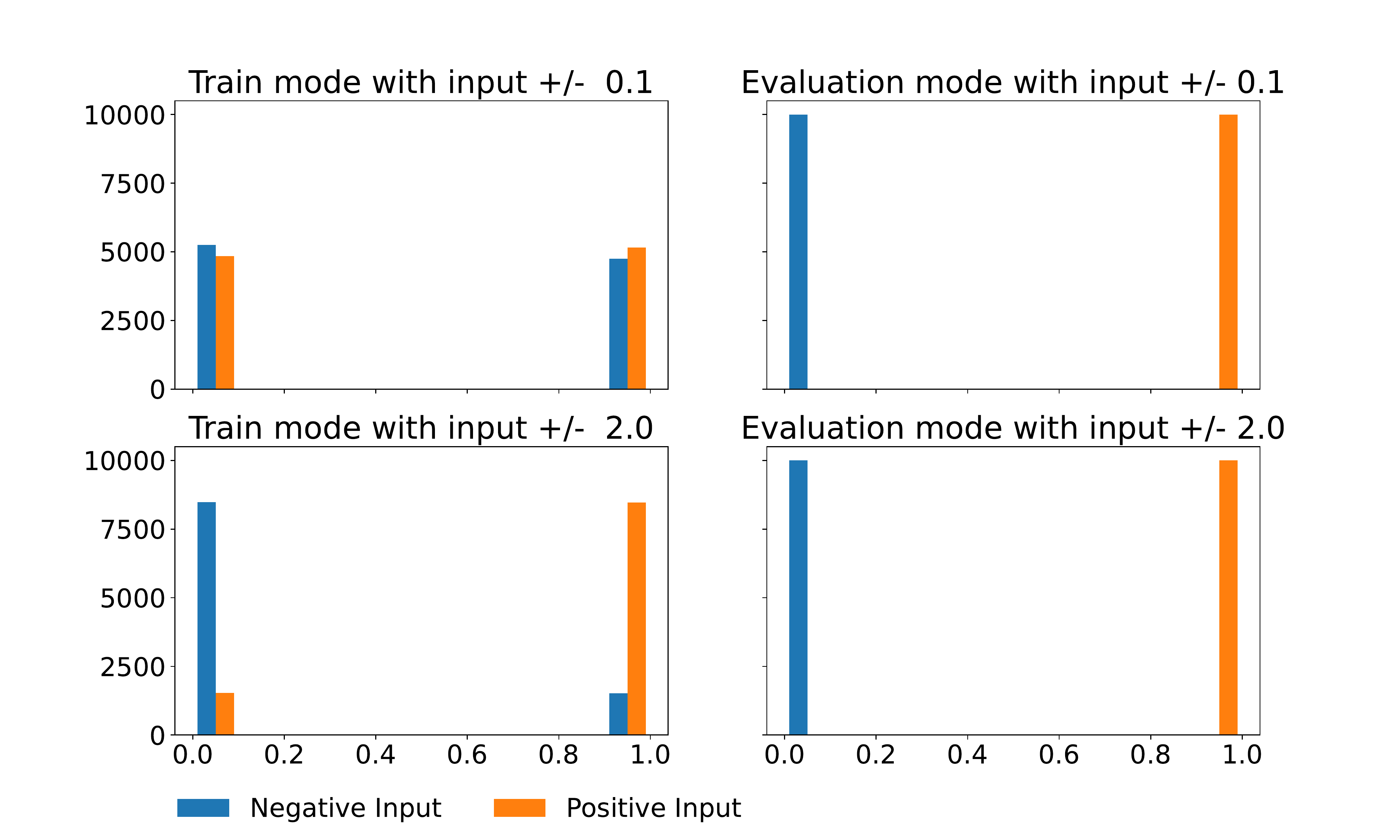}
        \caption{Histogram of the output of the ST-DRU when calculating the output 10k times for different input values (positive and negative) for both train and evaluation mode}
        \label{fig:ST-DRU_response}
        \Description[Figure 4. Histogram of the output of the Discretize Regularize Unit adapted with a Straight Through Estimator]{The figure shows four histograms. On each histogram, we see the output of the Discretize Regularize Unit adapted with a Straight Through Estimator for 10000 times a positive input and 10000 times a negative input. On the left we see the output when the Discretize Regularize Unit adapted with a Straight Through Estimator is in training mode and on the right we see the output when the Discretize Regularize Unit adapted with a Straight Through Estimator is in evaluation mode. The top left histogram shows the outputs of the Discretize Regularize Unit adapted with a Straight Through Estimator in train mode for an input of plus and minus 0.1. We see that around 50\% of both the negative and positive input values results in an output of 0 and the remaining cases result in an output of 1. On the bottom left histogram we see the outputs for an input of plus and minus 2.0. Here, around 80\% of the negative inputs result in an output of 0, the remaining cases resulting in an output of 1. Similarly, around 80\% of the positive inputs result in an output of 1, the remaining cases resulting in an output of 0. On the right two histograms we see the output when the Discretize Regularize Unit adapted with a Straight Through Estimator is in evaluation mode. For both the input of 0.1 and the input of 2.0 we see that all of the negative inputs result in an output of 0 and all of the positive inputs result in an output of 1.}
    \end{minipage}
    \hfill
    \begin{minipage}[t]{0.49\linewidth}
        \centering
        \includegraphics[width=\histogramwidth\linewidth, trim=65 10 80 35, clip]{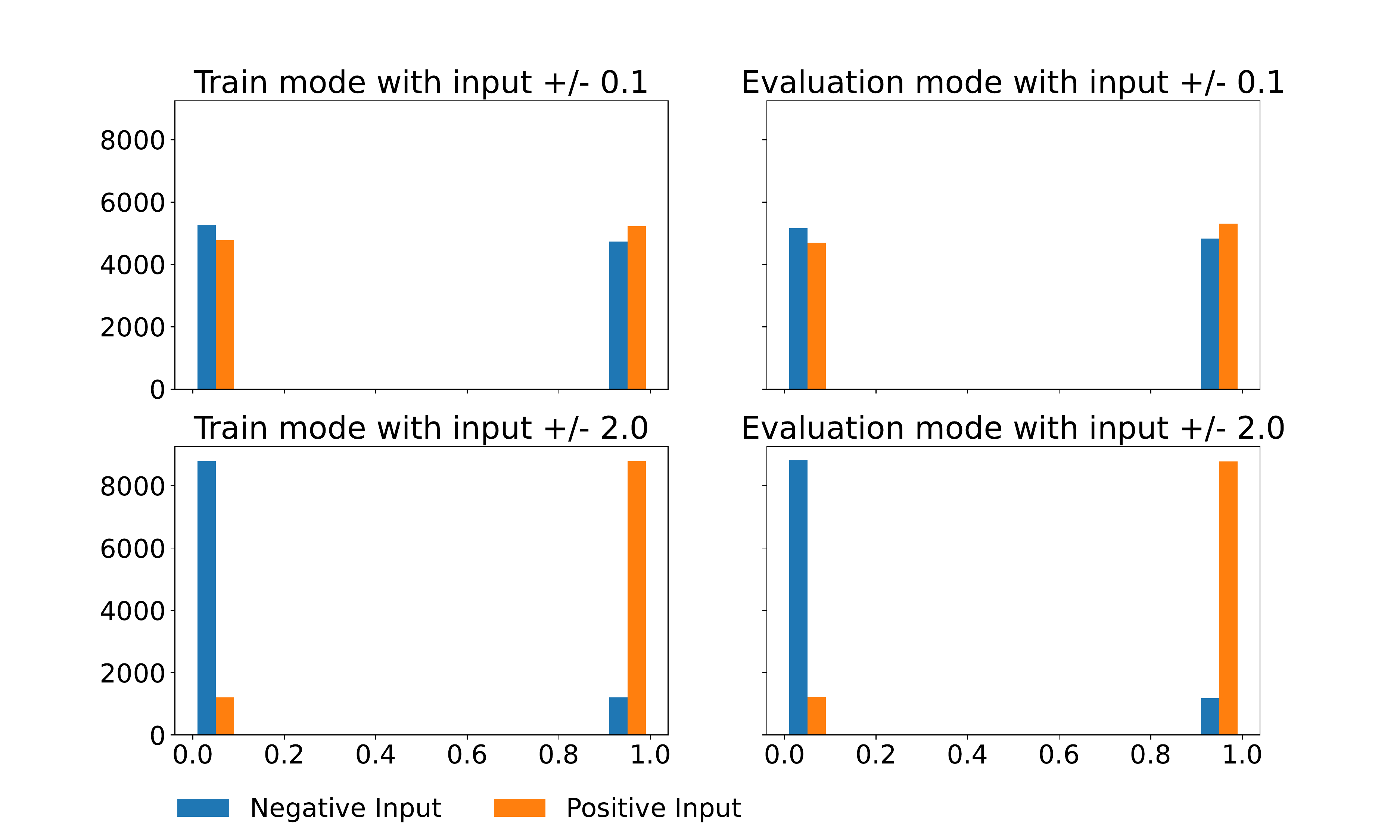}
        \caption{Histogram of the output of the ST-GS when calculating the output 10k times for different input values (positive and negative) for both train and evaluation mode}
        \label{fig:ST-GS_response}
        \Description[Figure 5. Histogram of the output of the Gumbel Softmax adapted with a Straight Through Estimator]{The figure shows four histograms. On each histogram, we see the output of the Gumbel Softmax adapted with a Straight Through Estimator for 10000 times a positive input and 10000 times a negative input. On the left we see the output when the Gumbel Softmax adapted with a Straight Through Estimator is in training mode and on the right we see the output when the Gumbel Softmax adapted with a Straight Through Estimator is in evaluation mode. The top left histogram shows the outputs of the Gumbel Softmax adapted with a Straight Through Estimator in train mode for an input of plus and minus 0.1. We see that around 50\% of both the negative and positive input values results in an output of 0 and the remaining cases result in an output of 1. On the bottom left histogram we see the outputs for an input of plus and minus 2.0. Here, around 85\% of the negative inputs result in an output of 0, the remaining cases resulting in an output of 1. Similarly, around 85\% of the positive inputs result in an output of 1, the remaining cases resulting in an output of 0. On the right two histograms we see the output when the Gumbel Softmax adapted with a Straight Through Estimator is in evaluation mode. For both the input of 0.1 and the input of 2.0 we see the same results as for the Gumbel Softmax adapted with a Straight Through Estimator in train mode.}
    \end{minipage}
\end{figure*}

\subsection{Gumbel Softmax (GS)}
The Gumbel Softmax (GS) \citep{jang2017categorical, maddison2017concrete} is a method to approximate a sample from a categorical distribution in a differentiable way. Normal sampling techniques are not differentiable and therefore not directly applicable in this context. The GS achieves this desirable property by using Gumbel noise and the gumbel-max trick\citep{gumbel1954statistical}. Using the gumbel-max trick, we can sample from a categorical distribution with class probabilities $\pi$ as described in Equation \ref{eq:gumbel_max_trick}.
\begin{equation}
    \begin{aligned}
        z = one\_hot(\underset{i}{argmax}[g_i + log \pi_i]) 
    \end{aligned}
    \label{eq:gumbel_max_trick}
\end{equation}
where $g_1, ... g_k$ are i.i.d. samples drawn from Gumbel(0,1), $i \in \{0, 1\}$, $\pi_0 = \sigma(x)$, $\pi_1 = (1 - \sigma(x))$, $x$ is the input of the discretization unit (output of the C-Net) and $\sigma(x)$ is the sigmoid function. To make this differentiable, we have to approximate the $one\_hot$ and $argmax$ functions with a $softmax$ function. Since we need two probabilities to obtain a categorical distribution for both states of a bit we will obtain the output message by using Equation \ref{eq:gumbel_softmax}.
\begin{equation}
    \begin{aligned}
        m = softmax \left(\frac{log(\sigma(x)) + g_1}{\tau}, \frac{log(1-\sigma(x)) + g_2}{\tau} \right)[0]
    \end{aligned}
    \label{eq:gumbel_softmax}
\end{equation}
where $x$ is the input of the discretization unit (output of the C-Net), $\tau$ is the softmax temperature and $\sigma(x)$ is the sigmoid function. A low temperature will result in an output that closely matches the output of Equation \ref{eq:gumbel_max_trick}. The higher the temperature the more the output will approach a uniform distribution. Figure \ref{fig:GS_response} shows the behaviour of the GS for different inputs in training and evaluation mode with temparature $\tau = 1.0$. 

\begin{figure*}
    \begin{minipage}[t]{\graphwidth\linewidth}
        \centering
        \includegraphics[width=\linewidth]{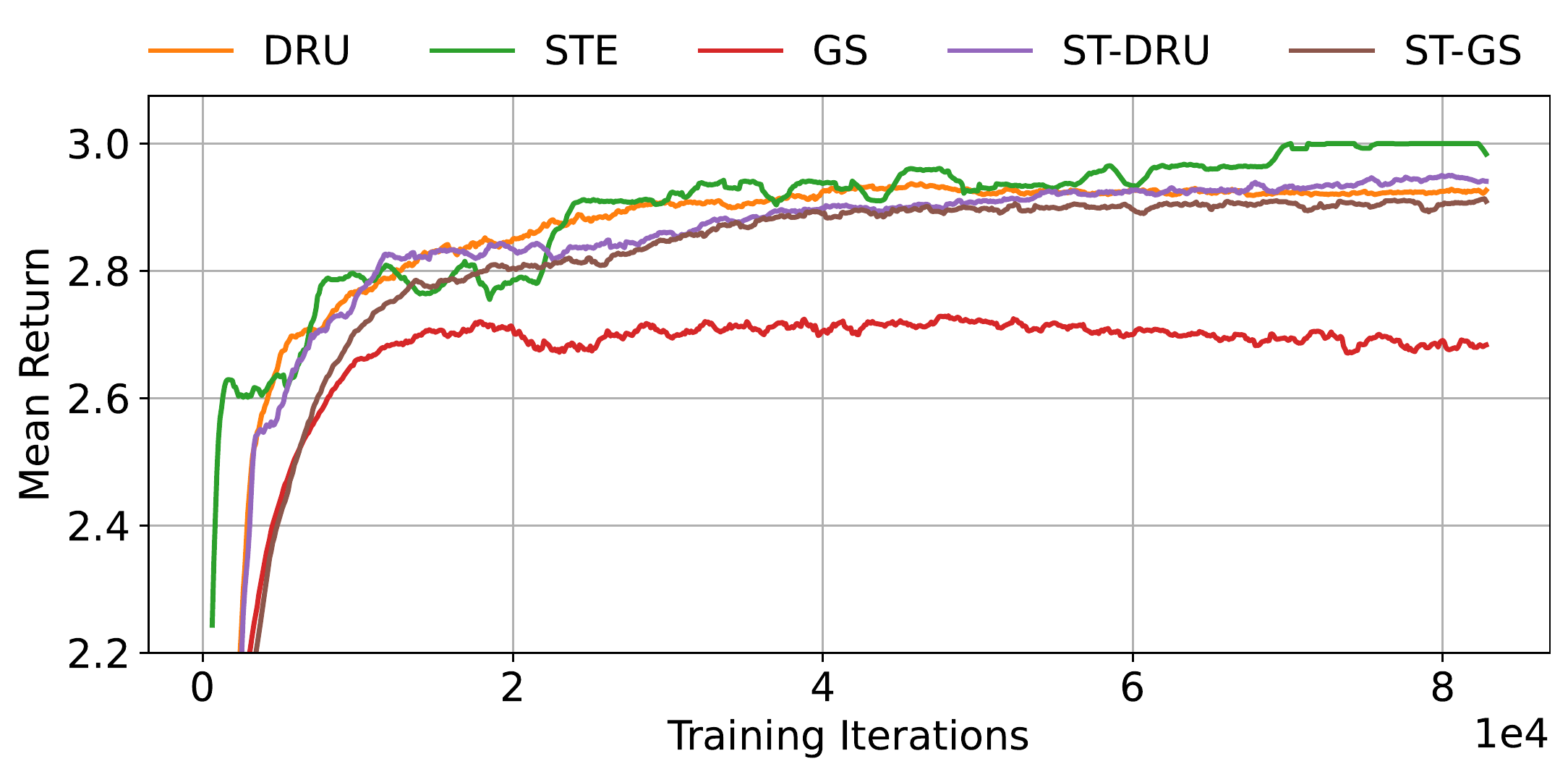}
        \caption{Results in the Simple Matrix environment}
        \label{fig:matrix_simple_eval_reward}
        \Description[Figure 6. Results in the Simple Matrix environment]{This figure shows the evolution of the mean return during training. We see that only the Straight Through Estimator method is able to achieve the maximum reward of 3 after 70k training iterations. The Discretize Regularize Unit, Straight Through Discretize Regularize Unit and the Straight Through Gumbel Softmax reach a reward of 2.9 after 40k training iterations. The Gumbel Softmax achieves the worst results. It reaches a reward of 2.7 after 20k training iterations.}
    \end{minipage}
    \hfill
    \begin{minipage}[t]{\graphwidth\linewidth}
        \centering
        \includegraphics[width=\linewidth]{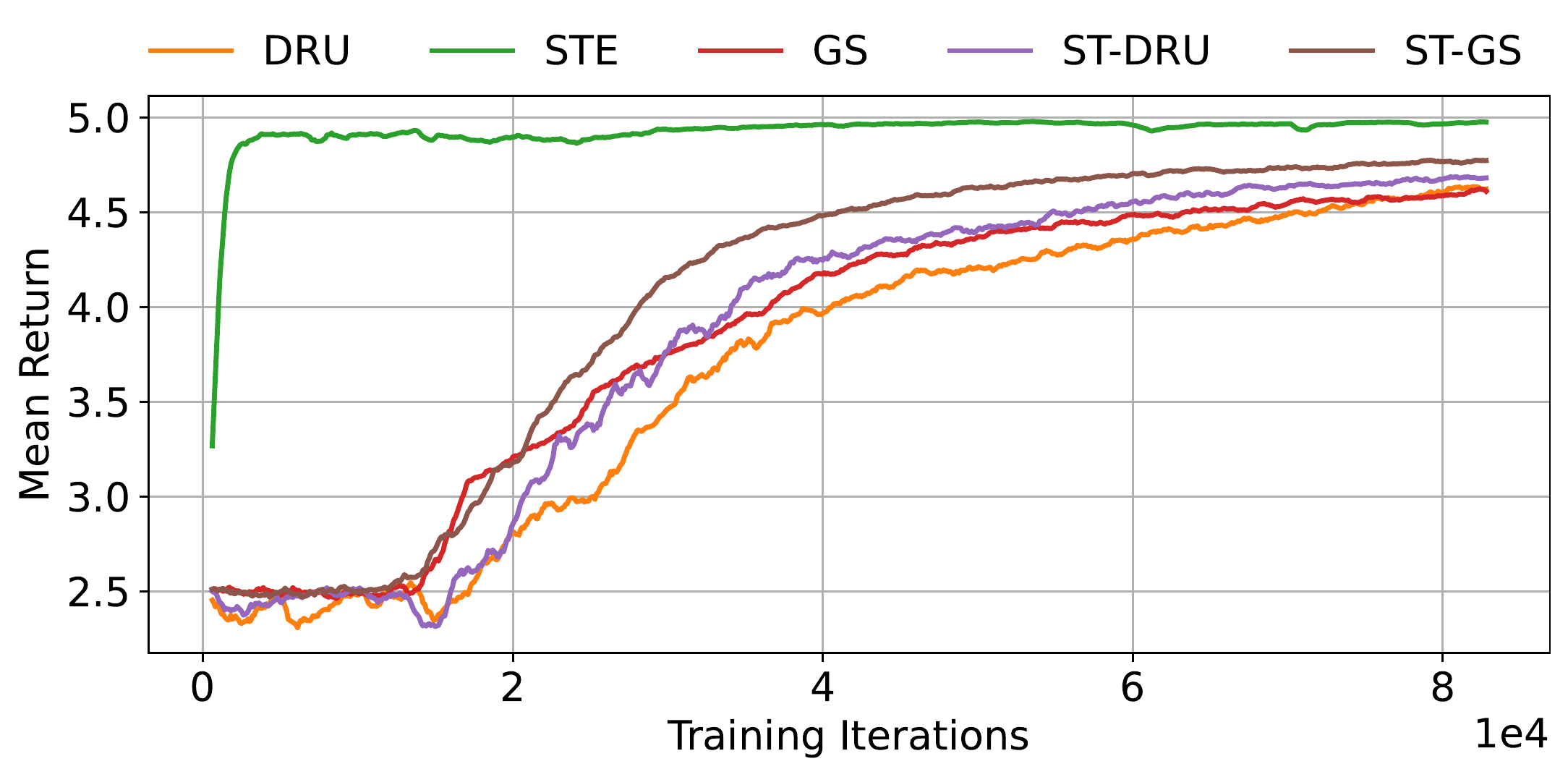}
        \caption{Results in the Complex Matrix environment}
        \label{fig:matrix_complex_eval_reward}
        \Description[Figure 7. Results in the Complex Matrix environment]{This figure shows the evolution of the mean return during training. We see that only the Straight Through Estimator method is able to achieve the maximum reward of 5 after 5k training iterations. The Discretize Regularize Unit, the Gumbel Softmax, the Straight Through Discretize Regularize Unit and the Straight Through Gumbel Softmax achieve a reward around 2.5 until they start improving after 15k training iterations. They reach a reward of 4.6 after 80k training iterations.}
    \end{minipage}
\end{figure*}

\subsection{ST-DRU}
We propose a novel discretization method (ST-DRU) that combines the DRU and STE methods. During execution, we discretize the input messages in the same way as mentioned in the DRU and STE methods, shown in Equation \ref{eq:DRU_discretize}. During training, we will use a different function for the forward and backward pass. In the forward pass, we add gaussian noise and apply the same discretization, resulting in Equation \ref{eq:STE_discretize}. 
\begin{equation}
    \begin{aligned}
        m = H(x + n)
    \end{aligned}
    \label{eq:STE_discretize}
\end{equation}
where $x$ is the input of the discretization unit (output of the C-Net) and $n$ is noise sampled from a Gaussian distribution with standard deviation $\sigma_G$. However, during backpropagation we use the gradients of Equation \ref{eq:DRU_regularize} instead. 
The advantage of this approach over the original DRU can be seen in Figure \ref{fig:ST-DRU_response}. The agents receiving the messages will receive binary messages from the start of training. The DRU uses continuous messages during training. Even though the agents are encouraged by the DRU to produce outputs with a high absolute value, it will take a while before the output messages will resemble binary messages. The ST-DRU will also encourage the agent to produce outputs that can easily be discretized. But, the receiving agents will immediately receive discrete messages, allowing them to learn to interpret them more quickly.

\subsection{ST-GS}
Similarly to the ST-DRU, we also test a straight through version of the GS as proposed by \citet{jang2017categorical}. Here, we will use the sampling technique described in Equation \ref{eq:gumbel_max_trick} to calculate the output, while using the gradients of the softmax approximation in Equation \ref{eq:gumbel_softmax} to train the communication network. Similarly to the DRU, the GS produces continuous messages during training. When evaluating the agents the GS will discretize the messages. If the sending agent is not producing outputs with a high enough absolute value, the difference between the messages at training time and at evaluation time will be very large. This prevents the receiving agents from correctly interpreting the message and choosing the appropriate actions. The ST-GS on the other hand will produce discretized messages at evaluation time and at training time, as can be seen in Figure \ref{fig:ST-GS_response}. This way, we make sure that the receiving agent knows how to correctly interpret discrete messages. 

\section{Experiments}
\label{sec:experiments}
In this section, we explain each of our experiments and analyze the results. For each experiment we show the average performance over five different runs for each of the discretization methods. The hyperparameters and network architectures are identical for each of the discretization methods since our hyperparameter search showed that the best hyperparameters and network architecture were not influenced by the choice in discretization method. All of our experiments are run using the RLlib framework \citep{liang2018rllib}. 

\subsection{Matrix Environment}
\label{sec:matrix_env}
The Matrix environment is inspired by the Matrix Communication Games presented by \citet{lowe2019measuring}. In the Matrix environment $N$ agents receive a natural number in $[0, M-1]$. The values for $N$ and $M$ can be chosen independent of each other. The agents are allowed to broadcast one message to the other agents before they have to indicate whether all agents received the same number or not. The odds of the agents receiving the same number are 50\% regardless of the value of $N$ and $M$. The minimum number of bits required to be able to represent each possible input number can easily be determined by applying a base two logarithm on $M$. The team reward in this environment is equal to the number of agents that correctly determined whether all agents got the same number or not. Therefore, the maximum reward is equal to $N$. Table \ref{tab:matrix_env} shows the reward matrices that correspond with a Matrix environment with $N = 2$ and any value for $M$.

\begin{table}[t]
    \caption{Reward matrices for the Matrix environment ($N = 2$ and any value for $M$). The agents have two possible actions, indicating they got the same number (S) or a different number (D)}
    \label{tab:matrix_env}
    \begin{subtable}[h]{0.45\linewidth}
        \centering
        \caption{The agents received the same number}
        \begin{tabular}{c|cc}
                & S     & D\\ \midrule
            S   & 2, 2  & 1, 1     \\
            D   & 1, 1  & 0, 0    
        \end{tabular}
    \end{subtable}
    \hfill
    \begin{subtable}[h]{0.45\linewidth}
        \centering
        \caption{The agents received a different number}
        \begin{tabular}{c|cc}
                & S     & D\\ \midrule
            S   & 0, 0  & 1, 1     \\
            D   & 1, 1  & 2, 2     
        \end{tabular}
    \end{subtable}
\end{table}

We examine the results for two different configurations of this environment which can be seen in Table \ref{table:matrix_configs}. In each of these experiments, we show the evaluation reward of our agents, measured by performing 100 evaluation episodes after each 100 training iterations. During the evaluation episodes, the agents do not explore and the discretization methods are applied in evaluation mode. In this environment, all of the agents are identical. Therefore, we can use parameter sharing between the agents, which improves their performance significantly as shown in the results of \citet{foerster2016learning}.

\begin{table}[t]
\centering
\caption{Different configurations of the Matrix environment}
\label{table:matrix_configs}
\begin{tabular}{ccc}
\toprule
                           & N & M \\ \midrule
Simple Matrix Environment  & 3 & 4 \\
Complex Matrix Environment & 5 & 256 \\
\bottomrule
\end{tabular}
\end{table}

\begin{figure*}
    \begin{minipage}[t]{\graphwidth\linewidth}
        \centering
        \includegraphics[width=\linewidth]{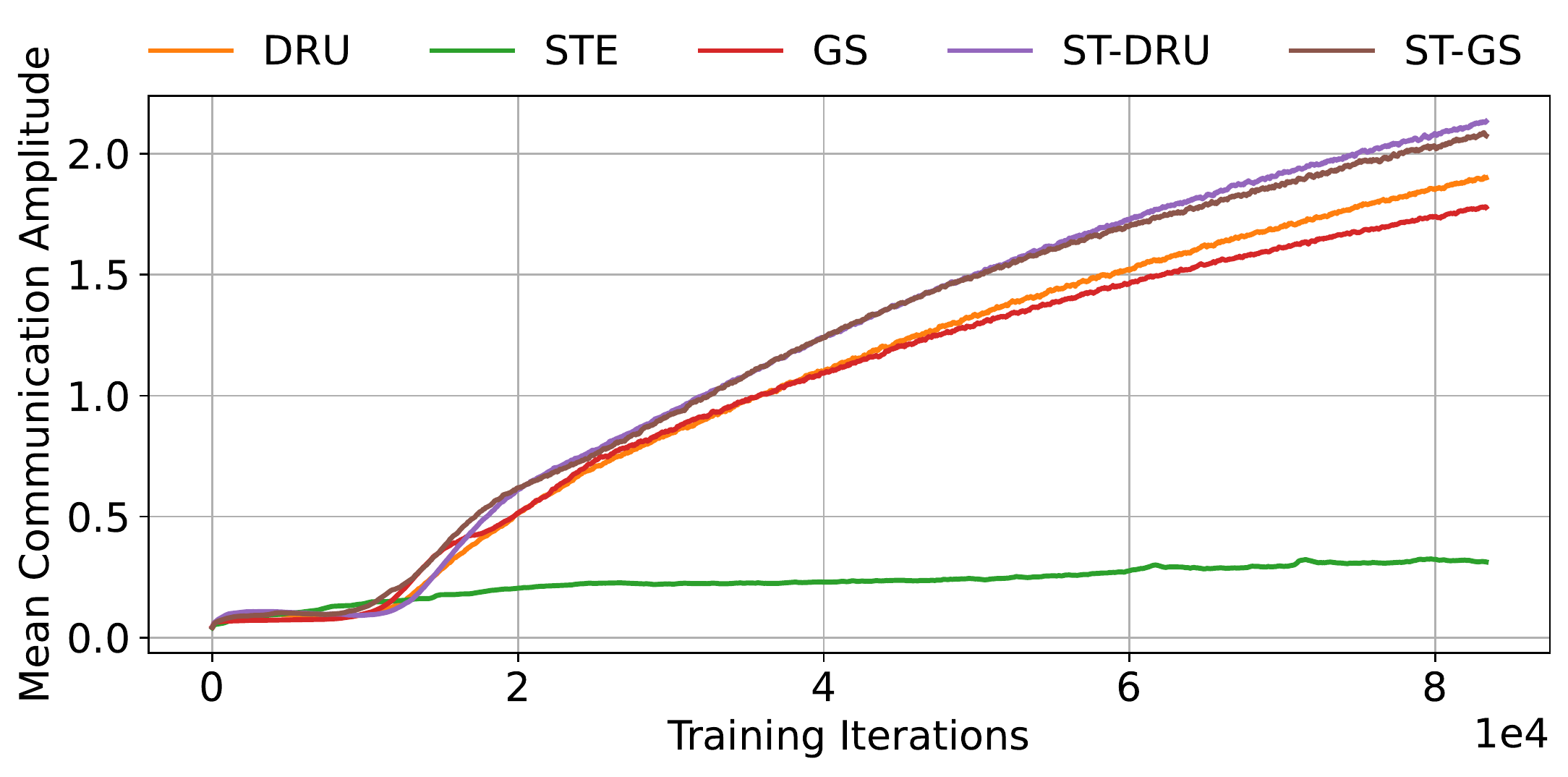}
        \caption{Communication amplitude in the Complex Matrix environment smoothed with an exponentially weighted moving average function with $\alpha = 5 \cdot 10^{-5}$}
        \label{fig:matrix_complex_comm_amplitude}
        \Description[Figure 8. Communication amplitude in the Complex Matrix environment]{This figure shows the evolution of the communication amplitude through training. We see that the curve of the Straight Through Estimator rises slowly throughout training, reaching a maximum value of around 0.4. The curves of the other methods starts rising quickly at around 15k training iterations. The Discretize Regularize Unit and the Gumbel Softmax reach a maximum value of around 1.8. The Discretize Regularize Unit adapted with a Straight Through Estimator and the Gumbel Softmax Straight Through Estimator reach a maximum value of around 2.1.}
    \end{minipage}
    \hfill
    \begin{minipage}[t]{\graphwidth\linewidth}
        \centering
        \includegraphics[width=\linewidth]{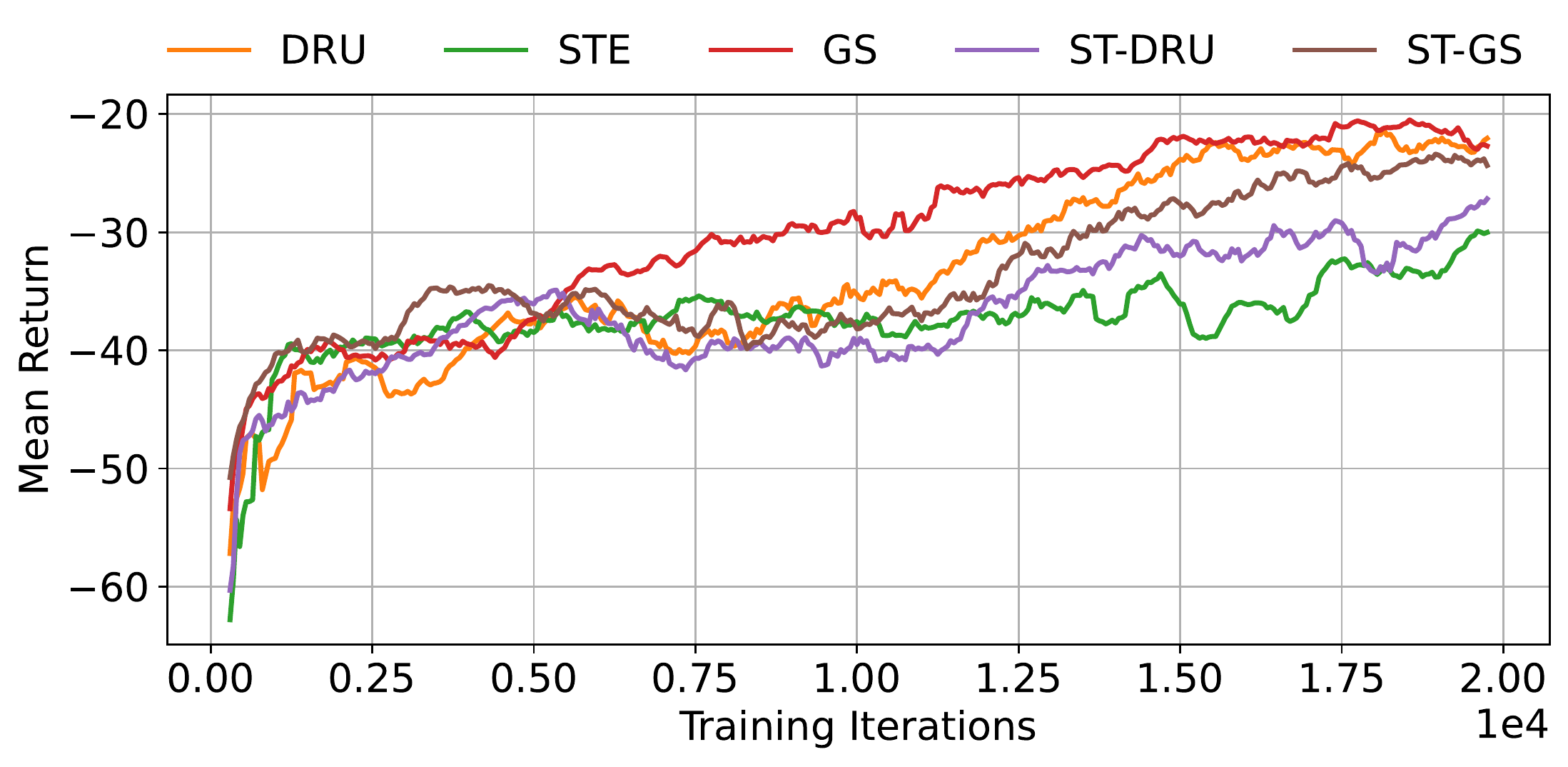}
        \caption{Results in the Speaker Listener environment}
        \label{fig:speaker_listener_eval_reward}
        \Description[Figure 9. Results in the Speaker Listener environment]{This figure shows the evolution of the mean return during training. We see that all of the methods reach a first plateau at a return of -40 after 1k training iterations. Afterwards the curve of the Gumbel Softmax keeps rising until reaching a return of -20 after 17.5k training iterations. The other methods remain at a return of -40 until 11k training iterations have been done. Then the curves of the Discretize Regularize Unit, the Straight Through Discretize Regularize Unit and the Straight Through Gumbel Softmax start rising as well. The Discretize Regularize Unit and the Straight Through Gumbel Softmax manage to reach a return of -25 after 17.5k training iterations. The Discretize Regularize Unit reaches a return of -27 and the Straight Through Estimator reaches a return of -30.}
    \end{minipage}
\end{figure*}

\begin{figure}[t]
    \centering
    \includegraphics[width=\graphwidth\textwidth]{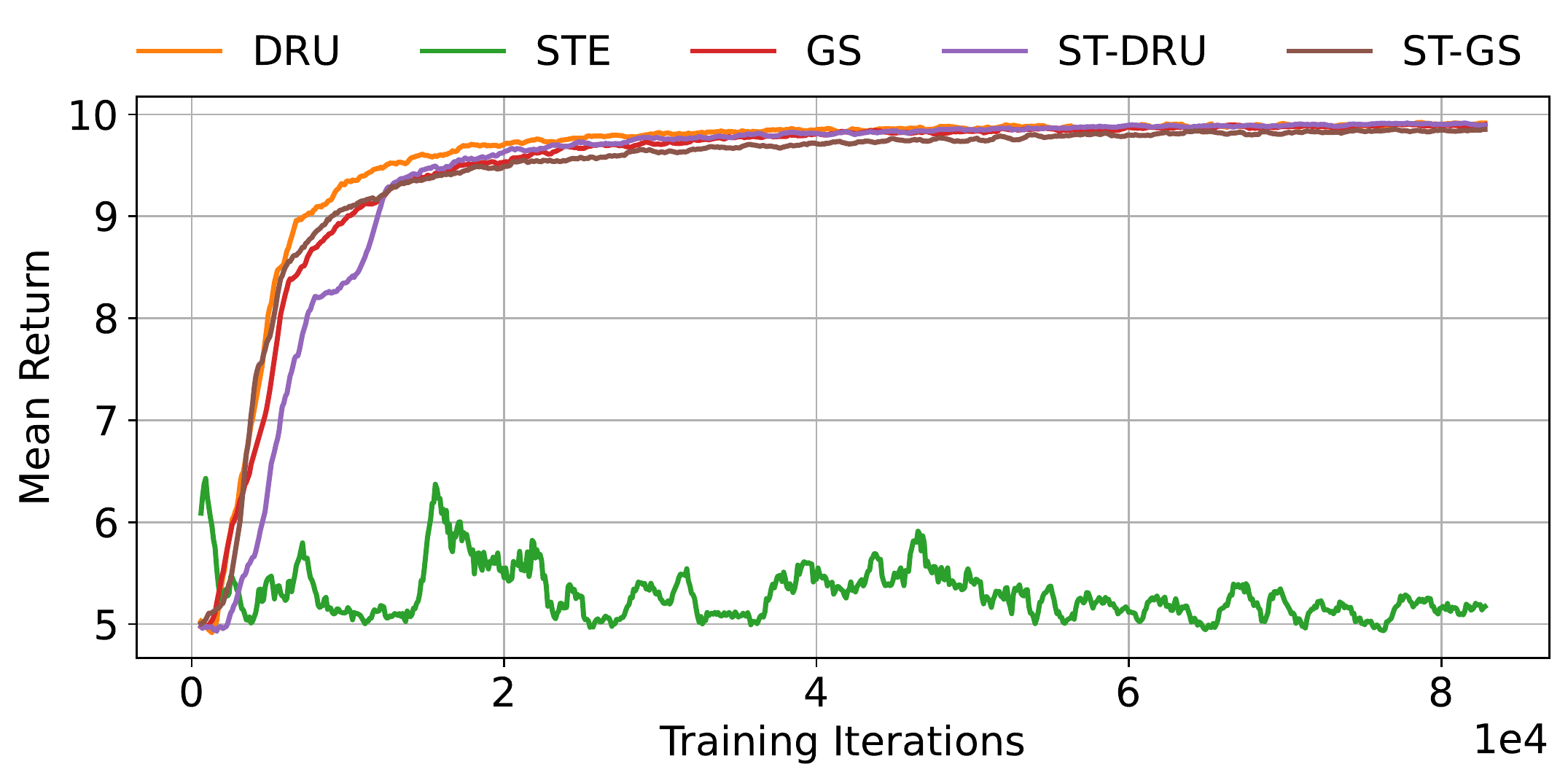}
    \caption{Results in the Matrix environment with a 50\% chance that a bit error will occur at a random location in the message}
    \label{fig:matrix_1_bitflip}
    \Description[Figure 10. Results in the Matrix environment with a 50\% chance that a bit error will occur]{This figure shows the evolution of the mean return during training. We see that only the Straight Through Estimator method is unable to learn in this environment. The Straight Through Estimator achieves a reward of 5 with peaks of 6.5. The other methods are able to reach a reward close to the maximum reward of 10. They reach their maximum return after 40k training iterations.}
\end{figure}

\subsubsection{Simple Matrix Environment}
In Figure \ref{fig:matrix_simple_eval_reward} and Table \ref{tab:conclusion_comparison}, the results for the different discretization methods are shown. The maximum reward the agents can achieve in this scenario is a reward of 3. We can see that most of the methods are able to achieve a reward very close to this maximum except for the GS. The ST-GS does not have the same issue as the GS to achieve the maximum reward. We also see that the STE method is faster at the beginning of the training but this difference disappears rather quickly. Due to the limited complexity of this environment, the differences between the methods are still small. 

\subsubsection{Complex Matrix Environment}
The Complex Matrix environment has more agents ($N = 5$) as well as more possible input numbers ($M = 256$). The agents need a message consisting of a full byte to be able to encode each of the possible input numbers. Figure \ref{fig:matrix_complex_eval_reward} and Table \ref{tab:conclusion_comparison} show the results of this experiment. The maximum reward in this configuration is 5. We see that the difference between the methods is larger than in the Simple Matrix environment due to the added complexity. We see that the STE method is the only one that is able to reach the maximum reward in this training period. It reaches a reward close to the maximum reward after only 5k training iterations. The other methods only start improving after 15k training iterations and take over 60k training iterations to reach their maximal performance. We can also see that the adapted versions of the DRU and GS which include the STE technique also perform better than the version without the STE technique. The ST-DRU has an average reward that is 0.072 higher than the DRU and the ST-GS has an average reward that is 0.176 higher than the GS during the final 10\% of training iterations. The communication amplitude in Figure \ref{fig:matrix_complex_comm_amplitude} provides an explanation for the training speed of the STE method. The communication amplitude is the mean absolute value of the input of the discretization unit. We can see a clear difference between the STE and the other methods. The communication amplitude of the STE stays below 0.5 while the communication amplitude of the DRU and GS approaches 1.8 and the communiation amplitude of the ST-DRU and the ST-GS exceeds 2.0. This is caused by the noise that is included in all of the discretization methods except for the STE. For a low communication amplitude the output of each of these discretization methods is still very random. The output during training will be determined by the noise instead of by the sign of the input which is done during evaluation. This encourages the agent to produce outputs with a higher communication amplitude. However, this puts a delay on the speed at which the agents can discover communication protocols. In Figure \ref{fig:matrix_complex_comm_amplitude} we see that the communication amplitude starts rising more quickly at around 15k training iterations. Once the communication amplitude starts rising we can see in Figure \ref{fig:matrix_complex_eval_reward} that the rewards that the agents receive also starts rising, indicating that they are starting to learn how to communicate with each other. 

\subsection{Speaker Listener Environment}

As a more complex environment, we use the speaker listener scenario from the particle environment by OpenAI \citep{lowe2020multiagent, mordatch2018}. This is one of the environments that was used to evaluate MADDPG \citep{lowe2020multiagent, mordatch2018}. In this environment there are two agents and three landmarks. One of the agents, the speaker, observes which landmark is the target during this episode. The speaker then has to communicate this information to the other agent, the listener. Next, the listener has to navigate to the target landmark. Both agents are rewarded using a team reward that is composed based on the distance of the listener to the target landmark. Contrary to the Matrix environment, the agents are not the same in this environment. The speaker will only consist of a communication policy while the listener will only consist of an action policy. In this experiment, we show the evaluation reward of our agents, measured by performing 10 evaluation episodes after each 50 training iterations. During the evaluation episodes, the agents do not explore and the discretization methods are applied in evaluation mode. 

Figure \ref{fig:speaker_listener_eval_reward} and Table \ref{tab:conclusion_comparison} show the results in the speaker listener environment. We see that the STE is no longer performing as good as in our earlier experiments. It results in the worst result while the GS and DRU provide the best results. The delay that is caused by the noise in the DRU, GS, ST-DRU and ST-GS is no longer the determining factor in the training speed. The exploration that is done in the communication policy by the DRU, GS, ST-DRU and ST-GS due to the noise appears to have a beneficial result.

\begin{table*}[t]
\centering
\caption{The average return and the standard deviation for each of the experiments during the final 10\% of training iterations}
\label{tab:conclusion_comparison}
\begin{tabularx}{\textwidth}{XXXXXX}
\toprule
Experiment Name     & DRU                  &  STE                 & GS                    & ST-DRU                   & ST-GS            \\
\midrule
Simple Matrix       & $2.924 \pm 0.152$    &  $2.999 \pm 0.005$   & $2.685 \pm 0.275$     & $2.944 \pm 0.112$         & $2.906 \pm 0.123$ \\
Complex Matrix      & $4.600 \pm 0.118$    &  $4.972 \pm 0.018$   & $4.588 \pm 0.048$     & $4.672 \pm 0.134$         & $4.764 \pm 0.047$ \\
Speaker Listener    & $-22.330 \pm 4.514$  &  $-32.224 \pm 6.833$ & $-21.533 \pm 4.758$   & $-29.966 \pm 9.430$       & $-24.211 \pm 5.177$\\
Error Correction    & $9.904 \pm 0.051$    &  $5.140 \pm 0.413$   & $9.892 \pm 0.051$     & $9.906 \pm 0.048$         & $9.842 \pm 0.059$  \\
\bottomrule
\end{tabularx}
\end{table*}

\subsection{Error Correction}

\begin{table}[t]
\centering
\begin{tabularx}{1\linewidth}{Xcc}
                                & \textbf{Before Errors}        & \textbf{After Errors} \\
\raisebox{0.6cm}{\textbf{DRU}}    & \includegraphics[width=\confusionwidth\linewidth, trim=30 90 45 100, clip]{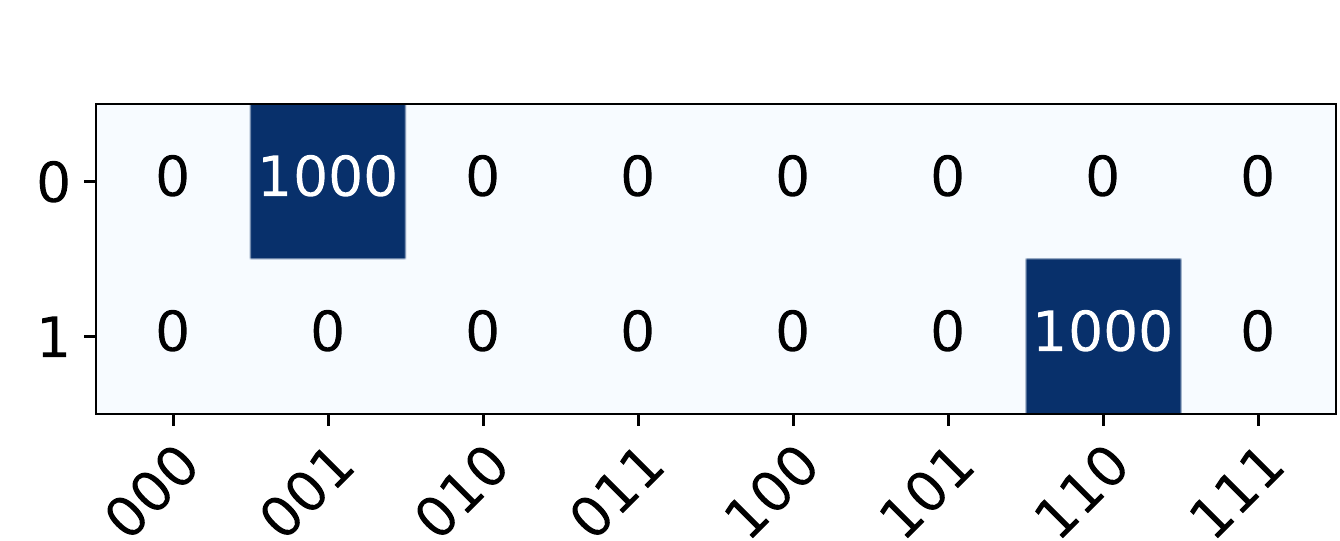}  &  \includegraphics[width=\confusionwidth\linewidth, trim=30 90 45 100, clip]{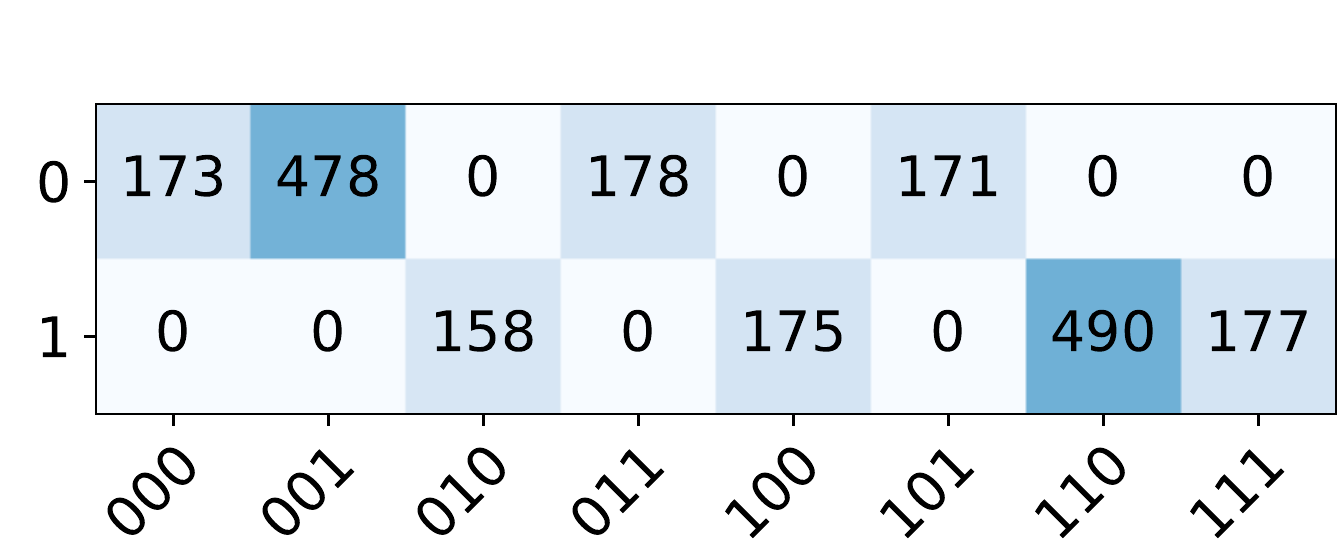}\\
\raisebox{0.6cm}{\textbf{STE}}    & \includegraphics[width=\confusionwidth\linewidth, trim=30 90 45 100, clip]{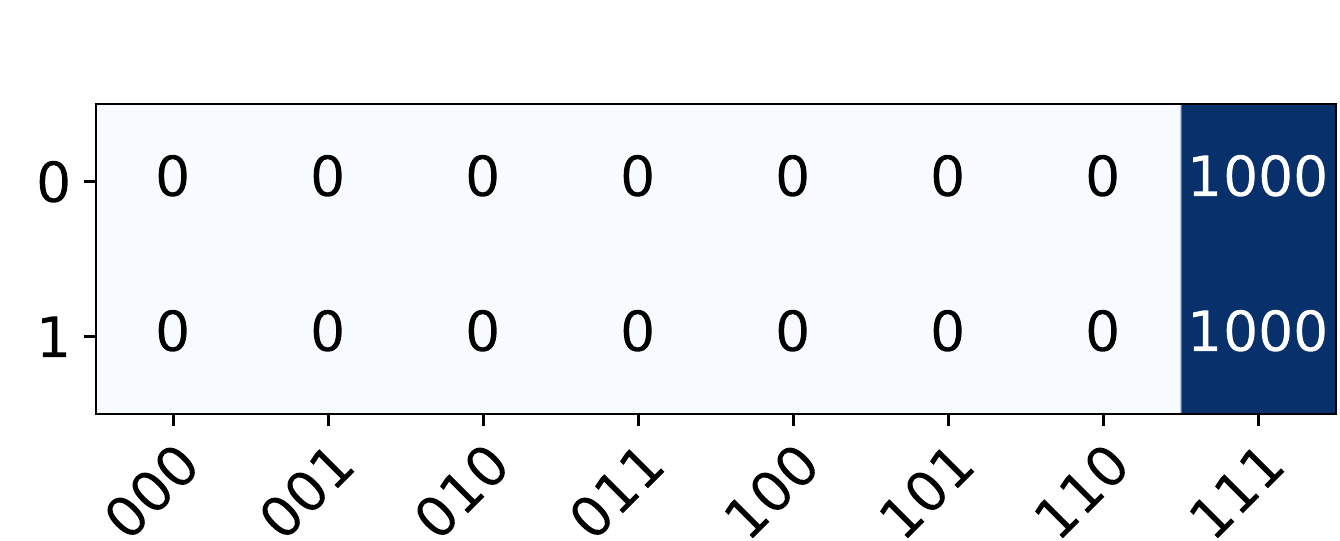}  &  \includegraphics[width=\confusionwidth\linewidth, trim=30 90 45 100, clip]{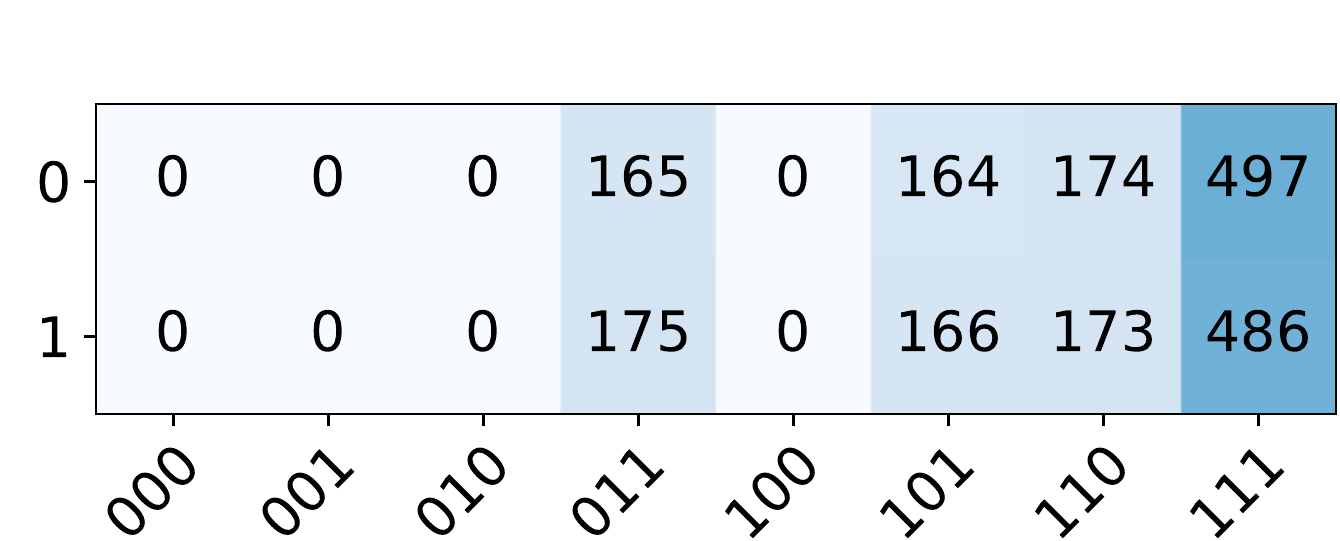}\\
\raisebox{0.6cm}{\textbf{GS}}     & \includegraphics[width=\confusionwidth\linewidth, trim=30 90 45 100, clip]{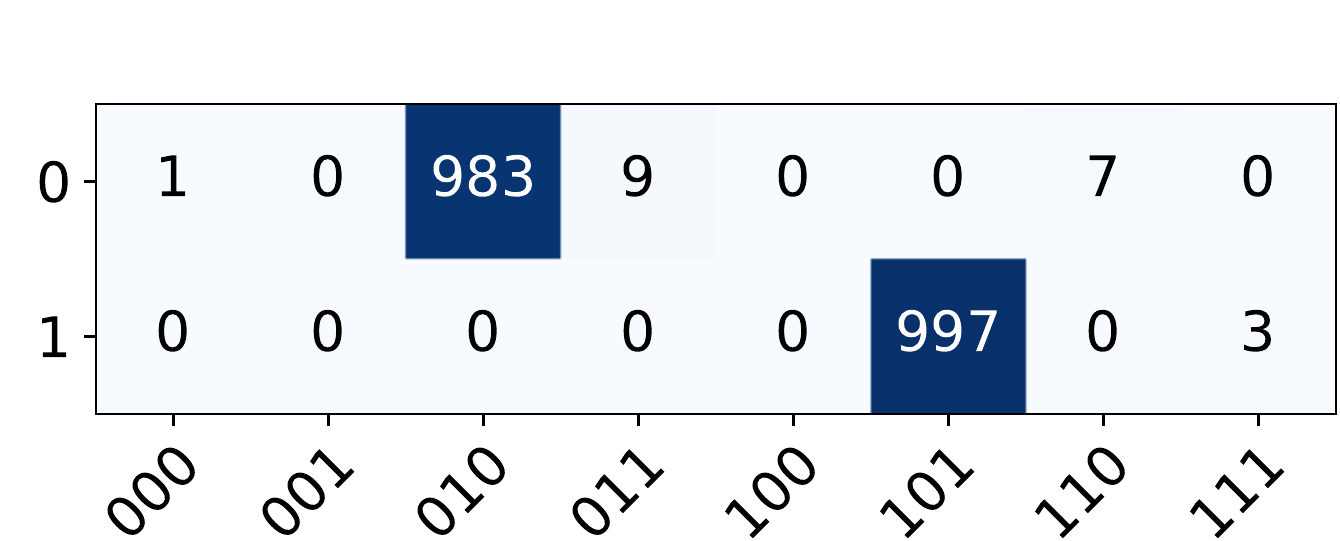}  &   \includegraphics[width=\confusionwidth\linewidth, trim=30 90 45 100, clip]{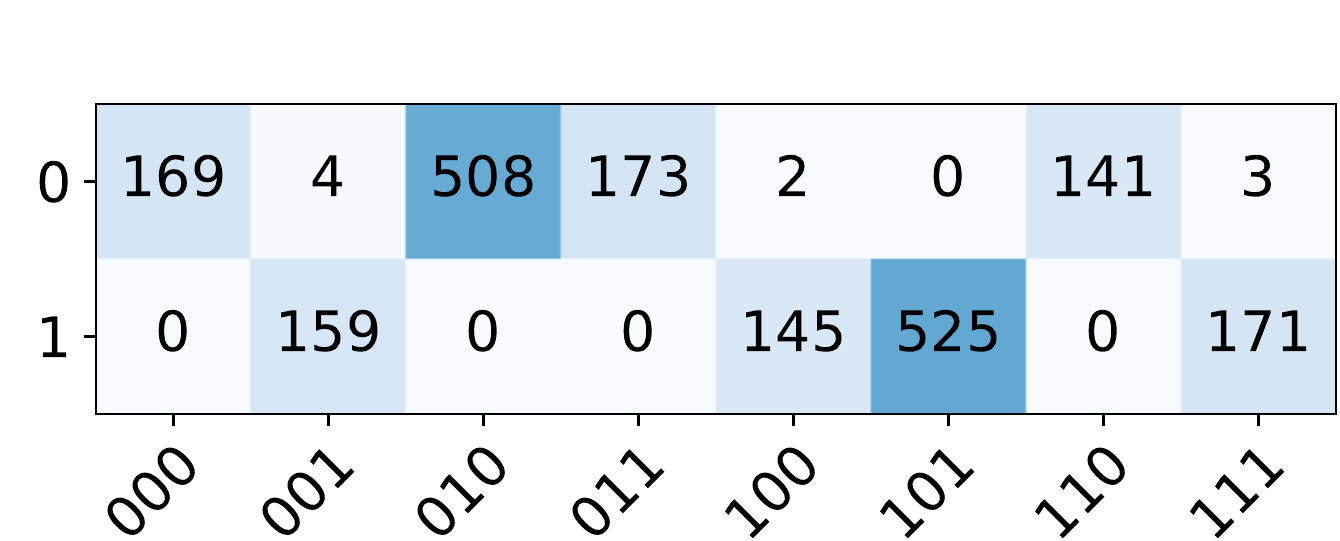}\\
\raisebox{0.6cm}{\textbf{ST-DRU}} & \includegraphics[width=\confusionwidth\linewidth, trim=30 90 45 100, clip]{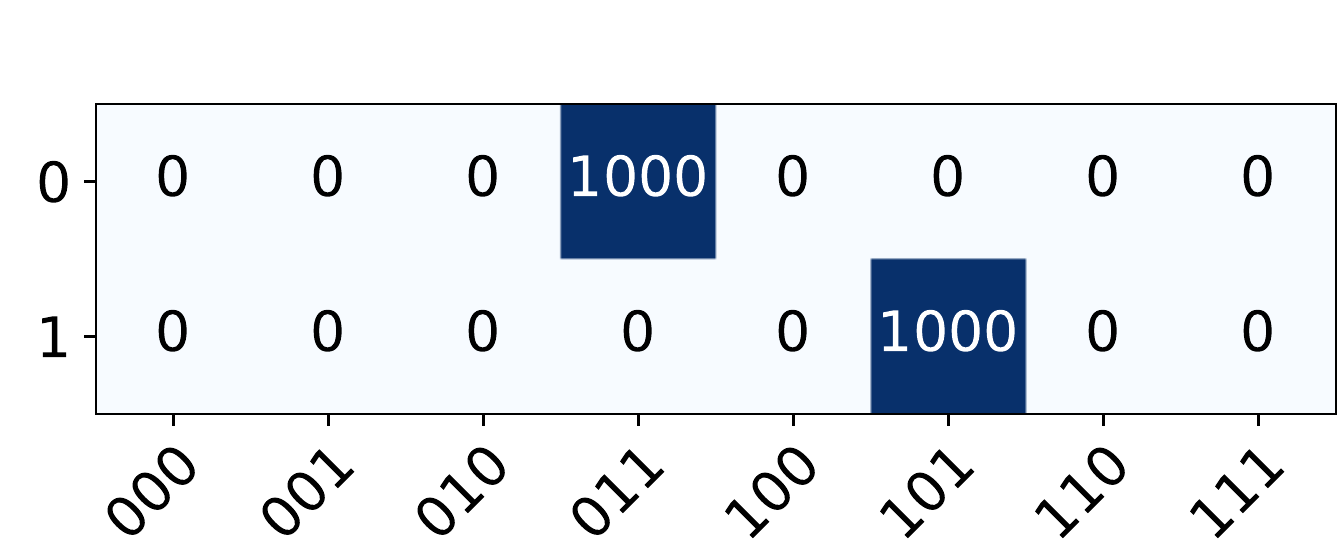} & \includegraphics[width=\confusionwidth\linewidth, trim=30 90 45 100, clip]{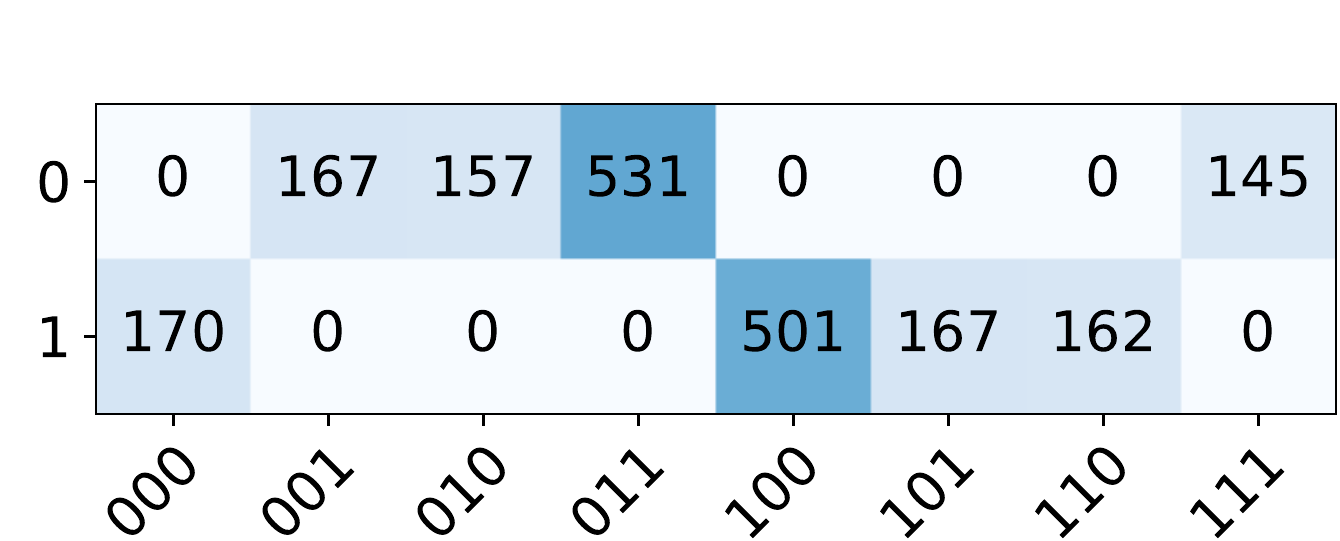} \\
\raisebox{0.6cm}{\textbf{ST-GS}}  & \includegraphics[width=\confusionwidth\linewidth, trim=30 90 45 100, clip]{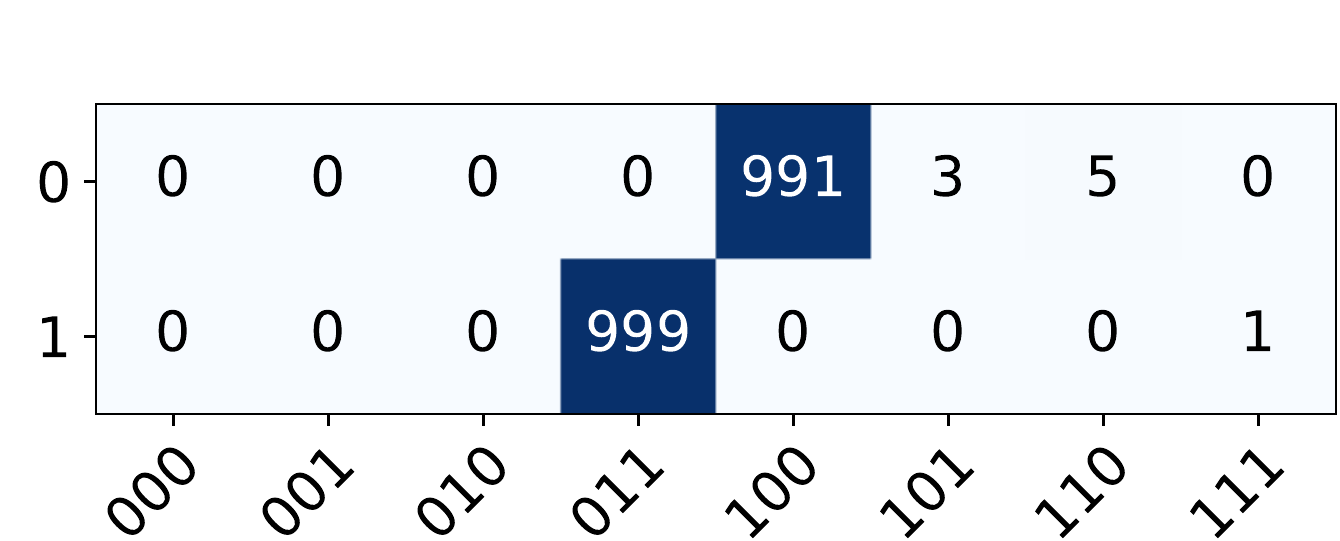}  & \includegraphics[width=\confusionwidth\linewidth, trim=30 90 45 100, clip]{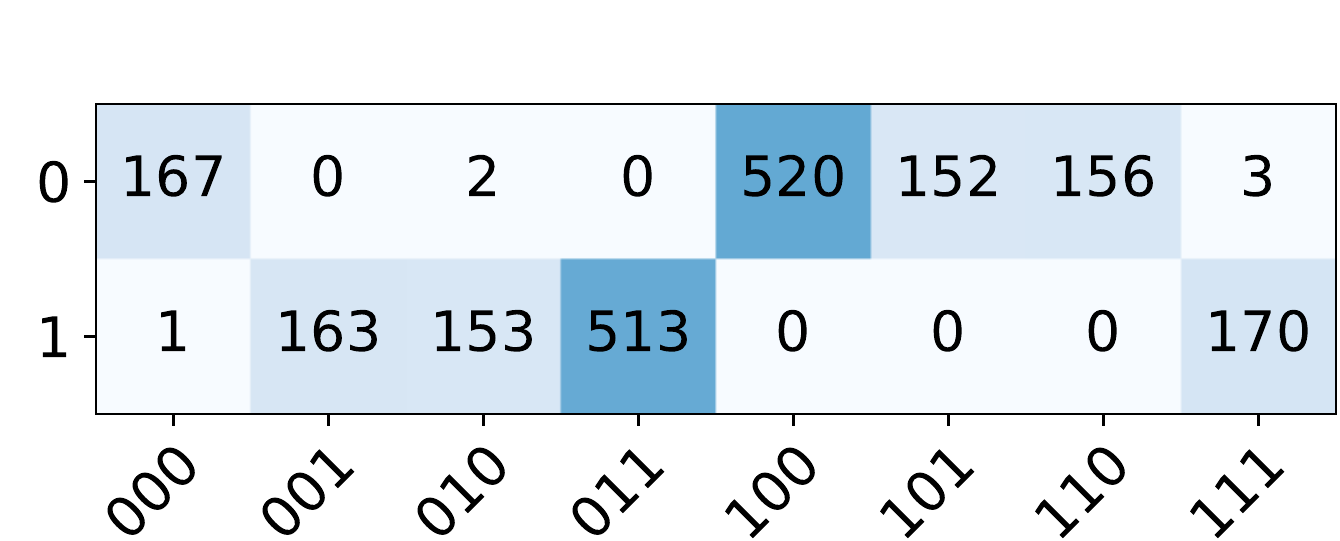}
\end{tabularx}
\captionof{figure}{The communication protocol for the Matrix environment ($N = 10$, $M = 2$) for each of the discretization units before and after introducing errors. On the y-axis the different possible input numbers are displayed. On the x-axis the different possible output messages are displayed.}
\label{fig:confusion_matrices}
\Description[Figure 11. The communication protocol for the Matrix environment before and after the introduction of errors]{This figure shows the communication protocol for each of the discretization methods before and after errors are introduced. For the Straight Through Estimator we see that for both possible input numbers the same message is generated. The Gumbel Softmax and the Straight Through Gumbel Softmax show a communication protocol where in almost all cases the same input is mapped on the same message. For the Gumbel Softmax 1\% of the inputs are mapped on different output messages. This causes an overlap between the messages for both input numbers after introducing errors in 0.5\% of the cases. The remaining cases result in an output message that is completely distinguishable from the other input number before and after introducing errors. For the Straight Through Gumbel Softmax 0.5\% of the input numbers are mapped onto different numbers. This causes an overlap between the messages for both input numbers after introducing errors in 0.3\% of the cases. The remaining cases result in an output message that is completely distinguishable from the other input number before and after introducing errors. The Discretize Regularize Unit and the Discretize Regularize Unit adapted with a Straight Through Estimator produce an communication protocol that always maps the same input number on the same message. After introducing errors, we see that the output messages do not show any overlap between the two possible input messages.}
\end{table}

In addition to comparing the different discretization methods in ideal circumstances, we also want to make this comparison in a situation with more uncertainty. Therefore, we perform some additional experiments on the Matrix environment as discussed in Section \ref{sec:matrix_env}. However, instead of perfect communication circumstances without any errors as done before, we flip a certain amount of random bits with a certain probability. This causes the receiver to receive different information than intended by the sender. Depending on the maximum amount of bits that can be flipped, the agents need more message bits to be able to counteract the errors that are introduced. We use a simple Matrix environment with $N = 10$ and $M = 2$. In this experiment, we show the evaluation reward of our agents, measured by performing 100 evaluation episodes after each 100 training iterations. During the evaluation episodes, the agents do not explore and the discretization methods are applied in evaluation mode. In this environment, all of the agents are identical. Therefore, we can use parameter sharing between the agents, which improves their performance significantly as shown in the results of \citet{foerster2016learning}. 

We perform a test where there is 50\% chance that an error will occur. Normally, the agents would be able to represent both possible incoming numbers using a single bit. However, if they have to be able to correct the errors that get introduced, they require three bits. The results of this experiment can be seen in Figure \ref{fig:matrix_1_bitflip} and Table \ref{tab:conclusion_comparison}. We see that the STE is not able to correct the errors that occur. Therefore, it is not able to achieve good results. However, the other discretization methods are able to detect and correct the errors. Our hypothesis is that these methods are more robust to errors due to the noise that is used within these discretization methods. 

To see how the agents are able to correct the introduced errors, we examine which message the agents choose for which incoming number. Figure \ref{fig:confusion_matrices} shows the different communication policies for the different discretization methods. We can see the output message depending on which input number was given to the agent before and after the errors are introduced. We see that the agents with the DRU, GS, ST-DRU or ST-GS have chosen messages where the possible messages after the introduction of errors do not overlap between the possible input numbers. This way the agents make sure that the messages are still comprehensible, even if errors occur. For the GS we see that there are 9 messages that overlap with the output messages for a different input number after error introduction. The same thing can be observed for the ST-GS in 6 cases. We see that when we use the STE, the agents are not able to find this communication protocol. Even before the errors are introduced, the messages for both possible inputs are the same. This indicates that the agents did not find a useful communication protocol. 

\section{Discussion}

\label{sec:discussion}
In our experiments, we compared different discretization techniques in different environments where the agents need to learn a communication protocol to achieve the goal. In this section, we discuss some general trends that we saw accross the experiments. Table \ref{tab:conclusion_comparison} shows how each of the different methods performed in each experiment. It shows the average return and standard deviation during the last 10\% of training iterations. 
In our results, we saw only small differences in a simple environment. However, the differences become a lot more apparent when using a more complex environment. The STE method performs very well in both the Simple and Complex Matrix environment, while performing worst in the speaker listener and not being able to achieve the goal in the error correction task. This makes the use of the STE as a standard method not recommended, especially in environments where perfect communication cannot be guaranteed. Similarly, the GS either performs the best among the tested methods or the worst. The ST-GS has a more consistent performance than the regular GS. The DRU and the ST-DRU perform very similar except in the speaker listener environment. There, the DRU clearly outperforms the ST-DRU. 

Overall, we can state that in most cases, either the DRU, ST-DRU or ST-GS should be used to discretize communication. These methods provide consistent results across the experiments while the STE and GS might achieve a higher return or be faster in some cases but fail dramatically in others. 

\section{Conclusion}
\label{sec:conclusion}

In this paper we compared several discretization methods in different environments with different complexities and challenges. We focus on the situation where these discretization methods are used to discretize communication messages between agents that are learning to communicate with each other while acting in an environment.

The results showed that the choice of discretization method can have a big impact on the performance. Across all of the experiments, the DRU, ST-DRU and ST-GS performed best. The STE performs a lot better in the matrix environment in terms of speed and return. However, in the speaker listener environment, the STE performs the worst and in the error correction task, it fails to learn a communication protocol. The GS performs best among all of the methods in the evironments where the STE fails, but performs the worst in the Matrix environment. The DRU, ST-DRU and ST-GS show more consistent results, close to the best result, making them a better standard choice. However, sometimes it may prove useful to perform additional experiments to establish the best discretization method.



\begin{acks}
This work was supported by the Research Foundation Flanders (FWO) under Grant Number 1S12121N and Grant Number 1S94120N. We gratefully acknowledge the support of NVIDIA Corporation with the donation of the Titan Xp GPU used for this research.
\end{acks}



\bibliographystyle{ACM-Reference-Format} 
\bibliography{references}

\end{document}